\documentclass[letterpaper]{article} % DO NOT CHANGE THIS
\pdfoutput=1

\usepackage{algorithm}
\usepackage{algorithmic}
\usepackage[margin=1in]{geometry}
\usepackage{caption} 
\usepackage{amsmath,amssymb,amsfonts,enumitem,amsthm,natbib,physics}
\usepackage{xcolor}
\usepackage{graphicx}
\usepackage{newfloat}
\usepackage{listings}
\usepackage{hyperref}
\usepackage{thmtools, thm-restate}

\newcommand\tred[1]{#1}
\newcommand{\X}{\mathbf{X}}
\newcommand{\cost}{\text{cost}}
\newcommand{\R}{\mathbb{R}}
\newcommand{\M}{\mathcal{M}}
\newcommand{\bigO}{\mathcal{O}}
\newcommand{\plog}{\text{polylog}}

\newtheorem{lemma}{Lemma}
\newtheorem{theorem}{Theorem}
\newtheorem{definition}{Definition}

\newtheorem{corollary}{Corollary}

\newcommand\xbar{\bar{x}}
\newcommand\ybar{\bar{y}}
\newcommand\Cbar{\bar{C}}

\newcommand\xhat{\hat{x}}
\newcommand\yhat{\hat{y}}

\newcommand{\prob}{\textsc{PrivEC}}
\newcommand{\algo}{\textsc{peCluster}}

\newcommand{\Se}{S_\epsilon}
\newcommand{\Sie}{S^{(i)}_\epsilon}

\title{Contrastive  explainable clustering with differential privacy}

\author {
    Dung Nguyen~\thanks{These authors contributed equally. Correspondence to: Dung Nguyen (dungn@virginia.edu)}~\thanks{Department of Computer Science, and Biocomplexity Institute and Initiative, University of Virginia, USA} ,
    Ariel Vetzler~\footnotemark[1]~\thanks{Department of Computer Science, Bar-Ilan University, Israel} ,
    Sarit Kraus~\footnotemark[3] ,
    Anil Vullikanti~\footnotemark[2] 
}
\date{}

\begin{document}

\maketitle

\begin{abstract}
This paper presents a novel approach to Explainable AI (XAI) that combines contrastive explanations with differential privacy for clustering algorithms. Focusing on k-median and k-means problems, we calculate contrastive explanations as the utility difference between original clustering and clustering with a centroid fixed to a specific data point. This method provides personalized insights into centroid placement. Our key contribution is demonstrating that these differentially private explanations achieve essentially the same utility bounds as non-private explanations. Experiments across various datasets show that our approach offers meaningful, privacy-preserving, and individually relevant explanations without significantly compromising clustering utility. This work advances privacy-aware machine learning by balancing data protection, explanation quality, and personalization in clustering tasks.
\end{abstract}

\section{Introduction}
\label{sec:intro}

Different notions of clustering are fundamental primitives in several areas, including machine learning, data science, and operations research~\cite{reddy2018data}. $k$-means and $k$-median clustering remain among the most important and widely used approaches, as demonstrated by recent advances in explainability, privacy, fairness, and contrastive learning~\cite{ghadiri2021socially,moshkovitz2020explainable,zehtabi2024contrastive,newling2016fast}.
These problems often involve significant trade-offs between accessibility, resource allocation, and overall cost.
For example, in emergency response planning, authorities must decide the optimal locations for ambulance stations to minimize response times across a city. Residents might question why an ambulance station isn’t located closer to their neighborhood, especially if they feel imbalances in resource distribution. Similarly, in retail, customers might wonder why a new store is not placed near their area, despite being part of a high-demand demographic. Explainability in these contexts provides transparency into how and why certain decisions are made, addressing questions like: "Why was this location chosen instead of another?" This is particularly important in applications where the consequences of clustering decisions directly affect individuals or communities.
\cite{bobek2022enhancing,ofek2024explaining} 

Such questions fall within the area of Explainable AI, which is a rapidly growing and vast area of research~\cite{finkelstein2022explainable,sreedharan2020bridging,miller2019explanation,boggess2023explainable,schleibaum2024adesse}. 
We focus on post-hoc explanations, especially contrastive explanations, e.g.,~\cite{miller2019explanation,pozanco2022explaining}, which address ``why P instead of Q?'' questions.
For example, in warehouse optimization, contrastive explanations clarify why a specific location was chosen as a distribution center, considering constraints like storage capacity or demand~\cite{zehtabi2024contrastive}. These methods are widely applied in multi-agent systems, reinforcement learning, and contrastive analysis~\cite{boggess2023explainable,boggess2022toward,sreedharan2021using,madumal2020explainable,sreedharan2020bridging}. In reinforcement learning, they explain actions by highlighting trade-offs, such as long-term rewards or risks~\cite{van2018contrastive}. 

\noindent Following the approach introduced in~\cite{zehtabi2024contrastive,pozanco2022explaining}, we 
%Our key innovation is a method to 
explain clustering decisions by comparing the costs in two scenarios: $cost(S)$, the cost of an optimal clustering solution on the whole dataset $\X = \{x_1, \ldots, x_n\}$ and $cost(S^{(i)})$, the cost of a modified solution where we fix a center at a desired location requested by agent $x_i\in\X$. 
We explain the decision by showing $cost(S^{(i)}) - cost(S)$, i.e., how much the overall clustering cost increases when we force a center to be in a specific location.
A higher clustering cost indicates worse performance. This comparison reveals the trade-offs in clustering: optimizing for one specific location often leads to a higher overall cost, meaning a worse solution for everyone else. By examining the difference between the optimal clustering cost and the cost of the forced fixed clustering, we can understand why the algorithm chose certain locations for centers and not others. This approach helps people grasp the complex balancing act involved in clustering decisions, especially when trying to distribute resources fairly \textcolor{black}{\cite{zehtabi2024contrastive}. }
% \textcolor{red}{
% In clustering problems, one can provide a "contrastive explanation" to agent $x_i$ by releasing the difference between $cost(S)$—the cost of the optimal clustering on the whole dataset—and $cost(S^{(i)})$—the cost of the optimal clustering with a centroid fixed at the desired location requested by agent $x_i$. Here, $x_i$ represents an individual data point from the dataset $\X = \{x_1, \ldots, x_n\}$, where $\X$ is the collection of all data points. Each $x_i$ corresponds to a specific agent.
% The difference implies the excessive cost of constructing a solution that selectively favors agent $i$, and serves as an explanation why we do not select that solution.}\\

\noindent Data privacy is a crucial concern across various fields, and Differential Privacy (DP) is one of the most widely used and rigorous models for privacy~\cite{dwork_and_roth}.
We focus on the setting where the set of data points $\X = \{x_1,\ldots, x_n\}$ are private; for instance, in the ambulance center deployment problem, each $x_i$ represents an individual requiring emergency services and seeking to keep their information private.
There has been a lot of work on the design of  differentially private solutions to clustering problems such as $k$-median and $k$-means in such a privacy model~\cite{ghazi:neurips20,gupta2009differentially,DBLP:conf/icml/BalcanDLMZ17,stemmer2018differentially}. 

\noindent While there has been significant progress in various domains of differential privacy, the intersection of explainability and differential privacy still needs to be explored. 
In clustering problems, building on the formalization of explanations for combinatorial problems 
we provide a private contrastive explanation to agent $x_i$ by computing $\cost(S^{(i)}_{\epsilon}) - \cost(S_{\epsilon})$. 
Here, $S_{\epsilon}$ represents a private solution using the privacy budget $\epsilon$, while $S^{(i)}_{\epsilon}$ denotes a private solution, when we constrain one center to be at a location specified by agent $x_i$
(this can be any point of interest, not necessarily the agent's own location). 
This difference quantifies how the clustering cost changes when accommodating agent $i$'s position, offering a privacy-preserving explanation of the clustering decision.
% However, giving such a private contrastive explanation to each agent $i$ naively using a $(w,t)$-approximate private clustering would lead to $(w,t\sqrt{n}/\epsilon)$-approximation, even if we use advanced composition techniques for privacy. 
However, giving such a private contrastive explanation to each agent $i$ naively using a private clustering algorithm would require a high privacy budget due to composition, which impact the accuracy, and lead to misleading or uninformative results. 
%This poor approximation factor implies that the output of the clustering algorithm would be far from optimal, potentially providing misleading or uninformative results.
The central question of our research: \emph{is it possible to offer each user an informative private contrastive explanation with a limited overall privacy budget?} 

\noindent
\textbf{Our contributions.}\\
1. We introduce the \prob{} problem, designed to formalize private contrastive explanations to all agents in clustering using $k$-median and $k$-means objectives.\\
2. We present an $\epsilon$-DP mechanism, PrivateExplanations, which provides a contrastive explanation to each agent while ensuring the same utility bounds as private clustering in Euclidean spaces, offering personalized insights without compromising privacy or clustering quality.
We use the private coreset technique of~\cite{ghazi:neurips20}, which is an intermediate private data structure that preserves similar clustering costs as the original data. \\
3. We evaluate our methods on diverse datasets with varying distributions and feature dimensions. Our results demonstrate privacy-utility trade-offs comparable to private clustering, with low clustering errors even at reasonable privacy budgets, showcasing the effectiveness of our approach.
Our research stands out by seamlessly integrating differential privacy into contrastive explanations, maintaining the quality of explanations even under privacy constraints. 
This work bridges the gap between privacy and explainability, marking a significant advancement in privacy-aware machine learning.
A key technical contribution of our work is the derivation of rigorous bounds on the approximation factors for all contrastive explanations, ensuring their reliability and effectiveness.
Due to space limitations, we only present major technical details in the main paper.
We maintain a full, updated version of this paper with complete proofs and extended experimental results at~\cite{nguyen2024contrastiveexplainableclusteringdifferential}.
%Table~\ref{table:notation} in the appendix summarizes all notations and their implications.
%%% Local Variables:
%%% mode: latex
%%% TeX-master: "main"
%%% End:

\section{Related Work}
\label{sec:related}

Our work considers differential privacy for explainable AI in general (XAI) and Multi-agent explanations (XMASE) in particular, focusing on post-hoc contrastive explanations for clustering. 
We summarize some of the work directly related to our paper; additional discussion is presented in the Appendix, due to space limitations.
Extensive experiments presented in \cite{saifullah2022privacy} demonstrate non-negligible changes in
explanations of black-box ML models through the introduction of privacy.

\cite{nguyen2023xrand} considers feature-based explanations (e.g., SHAP) that can expose
the top important features that a black-box model focuses on. To prevent such expose 
they introduced a new concept of achieving local differential
privacy (LDP) in the explanations, and from that, they established
a defense, called XRAND, against such attacks. They showed that
their mechanism restricts the information that the adversary
can learn about the top important features while maintaining
the faithfulness of the explanations.

\cite{goethals2022privacy} study the security of contrastive explanations, and introduce the concept of the ``explanation linkage attack'', a potential vulnerability that arises when employing strategies to derive contrastive explanations. 
To address this concern, they put forth the notion of k-anonymous contrastive explanations.
% Furthermore, the study highlights the intricate balance between transparency, fairness, and privacy when incorporating k-anonymous explanations. 
As the degree of privacy constraints increases, a discernible trade-off comes into play: the quality of explanations and, consequently, transparency are compromised.

Closer to our application is the work of \cite{georgara2022privacy}, which investigates the privacy aspects of contrastive explanations in the context of team formation. 
They present a comprehensive framework that integrates team formation solutions with their corresponding explanations, while also addressing potential privacy concerns associated with these explanations. Additional evaluations are needed to determine the privacy of such heuristic-based methods.

There has been a lot of work on private clustering and facility location, starting with ~\cite{gupta2009differentially}, which was followed by a lot of work on other clustering problems in different privacy models, e.g.,~\cite{huang2018optimal,stemmer:soda20,stemmer2018differentially,nissim2018clustering,feldman2017coresets}.
\cite{gupta2009differentially} demonstrated that the additive error bound for points in a metric space involves an $O(\Delta k^2\log{n}/\epsilon)$ term, where $\Delta$ is the space's diameter. Consequently, all subsequent work, including ours, assumes points are restricted to a unit ball.

We note that our problem has not been considered in any prior work in the XAI or differential privacy literature.
The formulation we study here will likely be useful for other problems requiring private contrastive explanations.

\section{Preliminaries}
\label{sec:prelim}

Let $\X\subset \mathbb{R}^d$ denote a dataset consisting of $d$-dimensional points (referred as agents).
% $\Arielnote{Contained in the unit ball.}
% Not needed since we say unit ball in the k,p clustering below
We consider the notion of $(k, p)$-clustering, as defined by Definition~\ref{def:kpcluster}. 
\begin{definition}
\label{def:kpcluster}
($(k, p)$-Clustering~\cite{ghazi:neurips20}). 
Given $k\in\mathbb{N}$, and a multiset $\X = \{x_1,\ldots, x_n\}$ of points in the unit ball, a $(k, p)$-clustering is a set of $k$ centers $\{c_1,\ldots,c_k\}$ minimizing $\cost^p_{\X}(c_1,\ldots,c_k) = \sum_{i\in[n]} \min_{j\in[k]} \| x_i - c_j\|^p$.
\end{definition}
\noindent For $p=1$ and $p=2$, this corresponds to the $k$-median and $k$-means objectives, respectively.
We drop the subscript $\X$ and superscript $p$, when it is clear from the context, and refer to the cost of a feasible clustering solution $S$ by $\cost(S)$.
\begin{definition}
($(w, t)$-approximation). 
Given $k\in\mathbb{N}$, and a multiset $\X = \{x_1,\ldots, x_n\}$ of points in the unit ball, let $OPT^{p,k}_{\X} = \min_{c_1,\ldots,c_k\in\mathbb{R}^d} \\ \cost^p_{\X}(c_1,\ldots,c_k)$ denote the cost of an optimal $(k, p)$-clustering.
We say $c_1,\ldots,c_k$ is a $(w, t)$-- approximation to a $(k, p)$-optimal clustering if 
$\cost^p_{\X}(c_1,\ldots,c_k) \leq w\cdot OPT^{p,k}_{\X} +t$.
\label{def:wt-approx}
\end{definition}
\noindent Let $OPT$ denote the cost of the optimal $(k,p)$-clustering, $OPT_i$ denote the cost of the optimal $(k,p)$-clustering, with a center fixed at position $z_i$ (the location chosen by agent $i$) and the remaining $k$-1 centers are chosen to optimize the objective.
Let $w', w''$ denote the maximum approximation (w.r.t. $OPT$ and $OPT_i$ respectively) of non-private clustering algorithms. These factors will be specified in Sections~\ref{sec:k-median} and~\ref{sec:kmeans}.

\noindent A coreset (of some original set) is a set of points that, given any $k$ centers, the cost of clustering of the original set is ``roughly'' the same as that of the coreset~\cite{ghazi:neurips20}.

\begin{definition}
  \label{def:coreset}
  For $\gamma, t > 0, p \geq 1, k, d \in \mathbb{N}$, a set $X'$ is a $(p, k, \gamma, t)$-coreset of $X\subseteq \R^d$ if for every $C = \{c_1, \ldots, c_k\}\in R^d$, we have $(1-\gamma)cost_X^p(C) - t \leq cost_{X'}(C) \leq (1+\gamma)cost_X^p(C) + t$.
\end{definition}

\noindent
\textbf{Privacy model.}
We use the notion of differential privacy (DP), introduced in~\cite{dwork_and_roth}, which is a widely accepted formalization of privacy. 
A mechanism is DP if its output doesn't differ too much on ``neighboring'' datasets; this is formalized below.
% Differential privacy is defined in terms of datasets that differ by one individual, called neighboring datasets, and requires that the output of a mechanism is (approximately) indistinguishable when run on any two neighboring datasets. 
% Formally, it is defined as:
\begin{definition}
$\M:\mathcal{X}\to\mathcal{Y}$ is $(\epsilon,\delta)$-differentially private if for any neighboring datasets $X\sim X^\prime \in\mathcal{X}$ and $S\subseteq\mathcal{Y}$,
\begin{align}
    \Pr[\M(X)\in S]\le e^\epsilon\Pr[\M(X^\prime)\in S] + \delta.\nonumber
\end{align}
If $\delta=0$, we say $\M$ is $\epsilon$-differentially private.
\end{definition}
\noindent We assume that the data points in $\X$ (i.e., users) are private, and  say $\X, \X'$ are neighboring (denoted by $\X\sim \X'$) if they differ in one data point.
When a value is disclosed to an individual agent $i$, it is imperative to treat the remaining clients in $\X-\{i\}$ as private entities. 
\begin{definition}
A mechanism $\M$ is $\epsilon$-$i$-exclusion DP if, $\forall X, X': i\in X, i\in X', X\setminus\{i\} \sim X'\setminus\{i\}$, and for all $S\subseteq Range(\M)$:
  \begin{align*}
  \Pr[\M(X) \in S] \leq e^{\epsilon}\Pr[\M(X') \in S].
\end{align*}

\noindent We extend this to say that $\M$ is $\epsilon$-$Y$-exclusion DP if the above holds $\forall X, X': Y\subset X, Y \subset X', X\setminus Y \sim X' \setminus Y $.
\end{definition}
\noindent We now define the \prob{} problem for providing private contrastive explanations, where each agent $x_i$ seeks an explanation for a center fixed at a location of their choosing denoted by $z_i$.
\begin{definition}
\textbf{Private and Explainable Clustering problem (\prob)}
Given an instance $\X\subset \mathbb{R}^d$, clustering parameters $k, p$, and a contrastive set of points 
$Z\subset \mathbb{R}^d$, the goal is to output:\\
% \begin{itemize}
% \item 
\textbf{Private}: An $\epsilon$-DP clustering solution 
$S_{\epsilon}$   (available to all)\\
% \item
\textbf{Explainable}: For each agent $x_i\in X$, output $\cost(S^{(i)}_{\epsilon})$ - $\cost(S_{\epsilon})$.\\
$S^{(i)}_{\epsilon}$ is a private solution computed by the clustering algorithm with one centroid fixed at the position requested by agent $i$. 

We assume that $S^{(i)}_{\epsilon}$ is not revealed to any agent, but $\cost(S^{(i)}_{\epsilon})$ - $\cost(S_{\epsilon})$ is released to agent $i$ as contrastive explanation, which is $\epsilon$-$i$-exclusion DP.
\begin{lemma} 
  \label{lemma:solution-utility}
  With probability at least $1-\beta$, cost($S_\epsilon$) (clustering cost) is a $(w,t)$-approximation of $OPT$, where\footnote{We use the notation $\bigO_{p,\alpha}$ to explicitly ignore factors of $p,\alpha$}:
  \begin{align*}
    w &= w'(1+\alpha),\\
    t &= w'\bigO_{p, \alpha}\left((k/\beta)^{O_{p,\alpha}(1)}.\plog(n/\beta)/\epsilon\right)
  \end{align*}
\end{lemma}
\noindent$\alpha$ is the approximation parameter for the utility of clustering and explanations. $\beta$ is the failure probability of the utility guarantees of clustering and explanations.
  \begin{restatable}{lemma}{validExplanation} {\bf DP of fixed centroid yield additional cost.}
  \label{lemma:valid-explanation}
 % (Full proof in Lemma~\ref{lemma:valid-explanation-full})
  Fix an $i$. If $OPT_i \geq w''(1+\alpha)OPT + t^{(i)}$, then with probability at least $1-2\beta$, cost($S_\epsilon$) and cost($S^{(i)}_\epsilon$) computed by Algorithm \algo{} satisfies that $cost(S^{(i)}_\epsilon) > cost(S_\epsilon$).
\end{restatable}

% \Arielnote{We already explained this in the introduction, I think we need to remove it from the introduction.}

% \anil{we only state the problem informally in the intro, so that the contributions are easy to follow}

\end{definition}

\section{\algo{} Mechanism}

We design  \textsc{PrivateExplanation} (Algorithm~\ref{alg:Private-Clustering}) for providing contrastive explanations for each agent.
Specifically, it takes as inputs: 
(1) $x_i$ which specifies the location of each agent $i$, and the contrastive location $z_i$ for which they want an explanation,
(2) original and target dimensions (d, d'), number of clusters ($k$), privacy budget $\epsilon$, and $\zeta$ (explained later).
The algorithm's key components are:

\begin{algorithm}[ht]
  \caption{PrivateExplanation\\ 
    \textbf{Input:} $(x_1, \ldots, x_n), (z_1, \ldots, z_n), d, d', k, \epsilon, \zeta$ \\\
    \textbf{Output:} $(\epsilon,\delta)$-differentially private explanation for agents
  }
  \label{alg:Private-Clustering}
  \begin{algorithmic}[1]
    % \STATE $\zeta = 0.01\left(\frac{\alpha}{10\lambda_{p,\alpha/2}}\right)^{p/2}$
    \STATE $(x'_1, \ldots, x'_n)$ $\gets$ $DimReduction((x_1,\ldots, x_n), d, d')$
    \STATE $Z'$ $\gets$ $DimReduction((z_1,\ldots, z_n), d, d')$
    \STATE $Y \gets \textsc{PrivateCoreset}^{\epsilon/2}(x'_1, \ldots, x'_n;\zeta)$
    \STATE$(c'_1,\dots,c'_k),cost(S'_\epsilon)\gets NonPrivateApprox(Y, k)$
    % \STATE $cost(S_\epsilon) = \left(\frac{\log(n/\beta)}{0.01}\right)^{p/2} cost(S'_\epsilon)$
    \STATE $cost(S_\epsilon) = RevertDimValue * cost(S'_\epsilon)$
    \STATE $c\gets \textsc{DimReverse}^{\epsilon/2}((c'_1,\dots,c'_k),(x'_1,\dots,x'_n))$
    \FOR {$z'_i \in Z'$}
        \STATE$cost(S'^{(i)}_\epsilon)\gets NonPrivateApproxFC(Y, k, z'_i)$
        \STATE  $cost(S^{(i)}_\epsilon) = RevertDimValue * cost(S'^{(i)}_\epsilon)$
    \ENDFOR
    \STATE \textbf{return} ${cost(S^{(i)}_\epsilon) - cost(S_\epsilon) | i\in range(1,\ldots, |X|)}$
  \end{algorithmic}
\end{algorithm}
\begin{itemize}
\item 
\textbf{Dimension Reduction:} Using \textsc{DimReduction} from \cite{ghazi:neurips20}, we transform input data (in dimension $d$) to a lower-dimensional space $d'$. This reduction is crucial since our coreset algorithm is exponential in the dimension, but by reducing to logarithmic dimensions, it becomes polynomial-time.
\item 
% (\textsc{PrivateCoreset}) into polynomial time.
\textbf{Private Coreset:} We create a differentially private coreset $Y$ using \textsc{PrivateCoreset} from \cite{ghazi:neurips20}, ensuring $\epsilon/2$ differential privacy; the coreset is defined in Definition~\ref{def:coreset}.
\item 
\textbf{Clustering:} The coreset is clustered using a non-private approximation algorithm (\textsc{NonPrivateApprox}). We can use a non-private clustering algorithm here since the coreset itself is already private, and by the Post-Processing property of differential privacy, the final result remains private.
\item 
\textbf{Cost Scaling:} In line 5 of the algorithm, by multiplying by $\textsc{RevertDimValue} = (\log(n/\beta)/0.01)^{p/2}$ we scale the clustering cost ($cost(S'_\epsilon)$) (in the low-dimensional space) back to the original dimension (cost($S_\epsilon$)) as shown in \cite{makarychev2019performance}. This reversal is necessary because while we computed costs in reduced dimensions for efficiency, we need the final cost in the original dimensions for accuracy.
\item 
\textbf{Dimension Reverse:} Centroids are mapped back to the original space using \textsc{DimReverse}, maintaining $\epsilon$-differential privacy.
\item 
\textbf{Contrastive Explanations:} For each data point, we execute fixed-centroid clustering (NonPrivateApproxFC) on the coreset, constraining one centroid to a location chosen by the agent. This algorithm, detailed in Sections~\ref{sec:k-median} and~\ref{sec:kmeans}, is our key contribution as it modifies standard k-means and k-median algorithms to fix one centroid while maintaining their original utility bounds from literature, ensuring meaningful explanations. Without utility bounds, agents could challenge the validity of the explanation, arguing that the fixed centroid might degrade the clustering solution to an unacceptable extent. However, by guaranteeing the same utility bounds as the original algorithms, we ensure that the explanations are grounded in the vicinity of optimal clustering solutions, leaving no room for users to dispute the fairness or validity of the explanation. This alignment between explanation quality and clustering utility reinforces the trustworthiness of the algorithm and the insights it provides. After clustering, we apply \textsc{RevertDimValue} to transform the cost back to the original space (cost($S^{(i)}_\epsilon$)).
By combining lines 5 and 9 of the algorithm, we derive the output: cost($S^{(i)}_\epsilon$) - cost($S_\epsilon$) for each agent. This value captures the loss of optimality when fixating one centroid, quantifying how much the clustering quality degrades due to this constraint, serving as a contrastive explanation.
\end{itemize}

\begin{restatable}{theorem}{mainprivacy} {\bf DP of Explanation.}
\label{theorem:privacy}
%(Full proof in the Appendix theorem ~\ref{theorem:privacy-full}.)
The solution $(c_1, \ldots, c_k)$ and $cost(S_\epsilon)$ computed by Algorithm \algo{} are $\epsilon$-DP.
For all clients $i$ and $S^{(i)}_\epsilon$ computed by Algorithm \algo{} is $\epsilon$-$i$-exclusion DP.
\end{restatable}
\noindent \textbf{Privacy analysis}, as demonstrated in Theorem~\ref{theorem:privacy}, we establishes the privacy guarantees of \algo{}. $Y$ coreset is $\epsilon/2$-differentially private as an output of $\epsilon/2$-DP algorithm. Consequently, $(c'_1, \ldots, c'_k)$ and $cost(S_\epsilon)$ maintain $\epsilon/2$-DP status, under the Post-Processing property.\\
Applying $\textsc{DimReverse}^{\epsilon/2}$ to find the centers in the original space, $c = \{c_1, \ldots, c_k\}$ is $\epsilon$-DP by Composition theorem. For each $i$, cost($S^{(i)}_\epsilon$) is produced by Post-Processing of $Y$ with only $z'_i$, hence $cost(S^{(i)}_\epsilon$) satisfies $\epsilon$-$i$-exclusion-DP.\\
\noindent \underline{\textbf{Running Time Analysis. }}\\
Algorithm~\ref{alg:Private-Clustering} has a total runtime of $O((k/\beta)^{O_{p,\alpha}(1)}\text{poly}(nd))$, which is polynomial in the input size. The key components contributing to this complexity include \textsc{PrivateCoreset}, \textsc{DimReverse}, and instances of $(k,p)$-clustering with and without fixed centers. \\
\textsc{PrivateCoreset} runs in $O((k/\beta)^{O_{p,\alpha}(1)}\text{poly}(n))$ time, as it sets $d' = O(p^4\log(k/\beta))$ to satisfy the Dimension-Reduction Lemma (Appendix Section B) and uses Lemma 42 from~\cite{ghazi:neurips20}. \textsc{DimReverse}, which includes the \textsc{FindCenter} operation (detailed in the Appendix), has a time complexity of $O(\text{poly}(np))$ and is invoked $k$ times. Additionally, we execute one standard $(k,p)$-clustering and $|X|$ instances of $(k,p)$-clustering with a fixed center. Together, these steps ensure the algorithm's overall polynomial runtime. All symbols used in this analysis are defined in Table~\ref{table:notation} in the Appendix.
\begin{theorem}
\label{Complexity-theorem}
  Assume there exist polynomial-time algorithms for $(k,p)$-clustering and $(k,p)$-clustering with a fixed center.
  The total running time of Algorithm~\ref{alg:Private-Clustering} is $O((k/\beta)^{O_{p,\alpha}(1)}\text{poly}(nd))$.
\end{theorem}
\noindent This computational complexity demonstrates that our algorithm is efficient for large datasets, balancing the additional overhead of fixed-centroid clustering with practical runtimes. Theorem~\ref{Complexity-theorem} follows from the detailed steps, as \textsc{PrivateCoreset} and \textsc{FindCenter} contribute manageable computational overhead.
Finally, the algorithm integrates a critical utility analysis to ensure robust performance. In the following sections, we present rigorous upper bounds and specific constraints for $k$-means and $k$-median, illustrating the practicality and effectiveness of our approach.

\noindent\underline{\textbf{Utility Analysis.}}
\textsc{PrivateCoreset} uses parameters $\zeta$ (which is a function of $\alpha$) and privacy budget $\epsilon/2$, derived from \cite{ghazi:neurips20} and detailed in our Appendix. This algorithm produces a coreset $Y$ that ensures the clustering cost on $Y$ closely approximates the cost on the projected dataset $X'$. Specifically, the approximation is within a $(1+0.1\alpha)$ factor, plus an additive $\overset{\sim}{O}(\text{polylog}(n/\beta)/\epsilon)$ term. Then, by applying the Dimensional Reduction lemma (in the Appendix), which states that the cost of a specific clustering on $X'$ ($d'$-dimensional space) is under some constant factor of the same clustering on $X$ ($d$-dimensional space), we can bound the $cost({S_\epsilon}^{(i)})$ by its optimal clustering $OPT_i$.
We first state the approximation factor derived using~\cite{ghazi:neurips20}, since this is used in our analysis.
\begin{restatable}{theorem}{explUtility}{\bf Cost of explanations due to privacy.} 
\label{theorem:explanation-utility}
  Fix an agent $i$. With probability at least $1-\beta$, cost($\Sie$) computed by Algorithm \algo{} is a $(w, t)$-approximation of $OPT_i$, with
  % -the optimal $(k,p)$-clustering cost with a center fixed at position $x_i$ and the remaining $k$-1 centers are chosen to optimize the objective, in which:
  \begin{align*}
    w &= w''(1+\alpha)\\
    t &= w''\bigO_{p,\alpha}\left((k/\beta)^{O_{p,\alpha}(1)}.\plog(n/\beta)/\epsilon\right)
  \end{align*}
\end{restatable}
\noindent As \(\Sie\) results from a randomized mechanism, its cost is higher than \(S_\epsilon\)'s most of the time with high probability, ensuring a positive private explanation.

\noindent \underline{\textbf{Tight Approximation Ratios. }}
The most challenging aspect of our analysis is determining the precise approximation factor $w''$ for k-means and k-median in the context of fixed-centroid clustering. In the following sections, we will present modifications to well-known k-median and k-means algorithms, adapting them for fixed-centroid clustering scenarios. We will then demonstrate that these modified algorithms achieve the same tight approximation factors.
Formally, we show how the well-known utility bounds of k-means and k-median can be preserved while fixing one centroid to a requested location, ensuring the robustness of these algorithms under such constraints.
Corollary~\ref{cor:private-kmedian-utility} and Corollary~\ref{cor:private-kmeans-utility} will conclude this section by presenting the specific, tight approximation ratios ($w''$) achieved after applying our \textsc{NonPrivateApproxFC} algorithm. These corollaries will provide detailed confirmation of our algorithm's effectiveness in achieving these optimal approximation ratios within the constraints of differential privacy.
%%% Local Variables:
%%% mode: latex
%%% TeX-master: "main"
%%% End:

\subsection{\textsc{NonPrivateApproxFC} for $k$-median}
\label{sec:k-median}
We have developed a non-private fixed centroid clustering algorithm, which we call \textsc{NonPrivateApproxFC}. This algorithm is an adaptation of~\cite{charikar1999constant}.
In the following section, we will prove that our modified algorithm, which works with a fixed centroid (referred to as $z$), achieves an 8-approximation factor.
% We obtain our results by adapting~\cite{charikar1999constant} to work with a fixed centroid (Moving forward in our discussion, we will refer to the fixed centroid as $C$).
To grasp how we adapted the algorithm to suit our needs, it's essential to understand the symbols used in ~\cite{charikar1999constant}. In this section, we adopt the notation from~\cite{charikar1999constant} to avoid confusion with the symbols used in this paper, where $d$ and $d'$ denote the original and reduced dimensions, respectively. In their work, $d_j$ represents the demand at each location $j \in N$, serving as a weight that reflects the importance of the location. $N$ refers to the set of agents ${1, \ldots, n}$.\\
For the conventional k-median problem, each $d_j$ is initially set to 1 for all $j\in N$. The term $c_{ij}$ represents the cost of assigning any $i$ to $j$, $x_{ij}$ represents if location $j$ is assigned to center $i$ and $y_i$ indicates if the location i is selected as a center.\\
\noindent We assume the fixed center is one of the input data points $N$. ~\cite{charikar1999constant} demonstrates that the $k$-median problem can be formulated as an integer programming problem, and in order to adapt the algorithm we add a constraint in line 9 to treat $z$ as a fixed centroid. This modification allows the algorithm to account for the fixed centroid requirement. We then relax the integer program (IP) into a linear program (LP) and show that it preserves the same utility bound as the original algorithm.
By specifying that $y_z \geq 1$, we ensure that $y_z$ is designated as a centroid in our linear programming formulation. Throughout the solution process, $y_z$ remains fixed as a centroid.

\begin{align}
    \text{minimize } &\sum_{i,j\in N} d_jc_{ij}x_{ij} \\
    \text{s.t.} \sum_{i\in N} x_{ij} &= 1 \text{ for each $j\in N$}; 
    \sum_{j\in N} y_j = k \\
    x_{ij} &\leq y_i \text{ for each $i, j\in N$}\\
    x_{ij}, y_{i} &\geq 0 \text{ for each $i, j\in N$}\\
  y_{z}, x_{zz} &\geq 1 \text{ for a fixed $z\in N$}
\end{align}

\noindent Let $(\xbar, \ybar)$ be a feasible solution of the LP relaxation and let $\bar{C}_j = \sum_{i\in N}c_{ij}\bar{x}_{ij}$ for each $j\in N$ as the total (fractional) cost of  client $j$.\\
Throughout the three steps, we demonstrate that solving this linear program with the added constraint does not introduce any additional approximation factor. The program is solved with the same efficiency and accuracy as it would be without the fixed centroid constraint.\\
\textbf{The first step.} We group nearby locations by their demands without increasing the cost of a feasible solution $(\xbar, \ybar)$, such that locations with positive demands are relatively far from each other. By re-indexing, we get $\Cbar_{z} \leq \Cbar_1 \leq \Cbar_2 \leq\ldots \Cbar_n$.\\
We will show that it's always possible to position $\Cbar_{z}$  as the first element of the list, i.e., $\Cbar_{z}$ is equal to the minimum value of all 
$\Cbar_j$. Recall that: $\Cbar_{z} = \sum_{i\in N}c_{i{z}}\xbar_{i{z}} 
          =\sum_{i\in N, i\neq z}c_{i{z}}\xbar_{i{z}} + c_{zz}\xbar_{zz}
          =0,$
          since we know that $\sum_{i\in N}x_{i{z}} = 1, x_{zz}\geq 1$ and $c_{zz} = 0$.\\
          The remaining work of the first step follows~\cite{charikar1999constant}. We first set the modified demands ${d_j}' \leftarrow d_j$. For $j \in N$, moving all demand of location $j$ to a location $i < j$ s.t. $d'_i > 0$ and $c_{ij}\leq 4\Cbar_j$, i.e., transferring all $j$'s demand to a nearby location with existing positive demand. Demand shift occurs as follows: $d'_i \leftarrow d'_i + d'_j, d'_j \leftarrow 0$. Since we initialize ${d'_{z}} = d_{z} = 1$, and we never move its demands away, it follows that $d'_{z} > 0$. \\
Let $N'$ be the set of locations with positive demands $N' = \{j \in N, d'_j > 0\}$. A feasible solution to the original demands is also a feasible solution to the modified demands.
\noindent \begin{lemma}
  \label{lemma:location-nearby}
    Locations $i, j \in N'$ satisfy: $c_{ij} > 4\max(\Cbar_i, \Cbar_j)$.
\end{lemma}
\noindent \begin{proof}
The lemma follows the demands moving step (in the first step of the algorithm): for every $j$ to the right of $i$ (which means $\Cbar_j \geq \Cbar_i$) and within the distance of $\Cbar_j$ (that also covers all points within distance $\Cbar_i$), we move all demands of $j$ to $i$, hence $j$ will not appear in $N'$. 
\end{proof}

%By Lemma~\ref{lemma:location-nearby}, for any pair $i, j \in N'$: $c_{ij} > 4\max(\Cbar_i, \Cbar_j)$.

\begin{lemma}
    The cost of the fractional $(\bar{x}, \bar{y})$ for the input with modified demands is at most its cost for the original input.
\end{lemma}
\noindent \begin{proof}
The cost of the LP $\Cbar_{LP} = \sum_{j \in N}d_j\Cbar_{j}$ and $\Cbar'_{LP} = \sum_{j \in N}d'_j\Cbar_{j}$. Since we move the demands from $\Cbar_j$ to a location $i$ with lower cost $\Cbar_i \leq \Cbar_j$ the contribution of such moved demands in $\Cbar'$ is less than its contribution in $\Cbar$, it follows that $\Cbar'_{LP}\leq \Cbar_{LP}$.
\end{proof}
\noindent \textbf{The second step.} We analyze the problem with modified demands $d'$. We will group fractional centers from the solution $(\xbar, \ybar)$ to create a new solution $(x', y')$ with cost at most $2\Cbar_{LP}$ such that $y'_i = 0$ for each $i\notin N'$ and $y'_i \geq 1/2$ for each $i\in N'$. 
%\anil{only for set $N'$, not $N$}
We also ensure that $y'_{z} \geq 1/2$ in this step, i.e., ${z}$ will be a fractional center after this.

A solution is called $1/2$-restricted if $y_j \geq 1/2$ for any point $j \in N$ and $y_j = 0$ otherwise. This restriction balances the assignment of demand, ensuring that no single center dominates excessively. The concept of $1/2$-restricted solutions is used to create more equitable distributions of demand and is key to transitioning to a $\{1/2, 1\}$-integral solution.
The next lemma leverages this property:

% \anil{haven't yet defined $1/2$-restricted}

\begin{lemma}
    For any $1/2$-restricted solution $(x', y')$ there exists a $\{1/2, 1\}$-integral solution with no greater cost.
\end{lemma}
\begin{proof}
  The cost of the $\frac{1}{2}$-restricted solution (by Lemma~7 of~\cite{charikar1999constant}) is:
  \begin{align}
    C'_{LP} &= \sum_{j\in N'}d'_jc_{s(j)j} - \sum_{j\in N'}d'_jc_{s(j)j}y'_j,
  \end{align}

\noindent Let $s(j)$ be $j$'s closest neighbor location in $N'$, the first term above is independent of $y'$ and the minimum value of $y'_j$ is $1/2$. We now construct a $\{1/2, 1\}$-integral solution $(\xhat, \yhat)$ with no greater cost. Sort the location $j\in N', j\neq {z}$ in the decreasing order of the weight $d'_jc_{s(j)j}$ and put ${z}$ to the first of the sequence, set $\yhat_j = 1$ for the first $2k-n'$ 
locations and $\yhat_j = 1/2$ for the rest. By doing that, we minimize the cost by assigning heaviest weights $d'_jc_{s(j)j}$ to the maximum multiplier (i.e., $1$) while assigning lightest weights $d'_jc_{s(j)j}$ to the minimum multiplier (i.e., $1/2$) for each $j\in N', j\neq z$. Any feasible $1/2$-restricted solution must have $y'_{z} = 1$ to satisfy the constraint of ${z}$ so that the contribution of $\yhat_{z}$ is the same as its of $y'_{z}$. It follows that the cost of $(\xhat, \yhat)$ is no more than the cost of $(x', y')$.
\end{proof}

\textbf{The third step.} This step is similar to the part of Step~3 of~\cite{charikar1999constant} that converts a $\{1/2, 1\}$-integral solution to an integral solution with the cost increases at most by $2$. We note that there are two types of center $\yhat_j = 1/2$ and $\yhat_j = 1$, hence there are two different processes. All centers $j$ with $\yhat_j = 1$ are kept while more than half of centers $j$ with $\yhat_j = 1/2$ are removed. Since we show that $\yhat_{z} = 1$ in the previous step, $z$ is always chosen by this step and hence guarantees the constraint of $z$.

\begin{theorem}
  \label{theorem:kmedian-fixed-center} {\bf Approximation factor of fixed centroid k-median.}
  For the metric k-median problem, the algorithm above outputs an $8$-approximation solution.
\end{theorem}
\noindent \begin{proof}
  It is obvious that the optimal of the LP relaxation is the lower bound of the optimal of the integer program. 
While constructing an integer solution for the LP relaxation with the modified demands, ~\cite{charikar1999constant} states that there is a 1/2-restricted solution ($x'$, $y'$) which costs at most $2\bar{C}_{LP}$. And now the third step multiplies this cost by a factor of $2$, making the cost of the solution (to the LP) at most $4\bar{C}_{LP}$. Transforming the integer solution of the modified demands to a solution of the original input adds an additive cost of $4\bar{C}_{LP}$ by 
Lemma 4 of~\cite{charikar1999constant} and the Theorem follows.
\end{proof}

\noindent Having demonstrated that our modification of ~\cite{charikar1999constant} to execute fixed-centroid k-median instead of standard k-median yields an 8-factor approximation of the optimal solution, we can now proceed to prove that our private explanation closely approximates the optimal solution for the fixed-centroid scenario.

\begin{corollary}
  \label{cor:Private-Clustering}
\label{cor:private-kmedian-utility}
  Running \algo~with \textsc{NonPrivateApproxFC} be the above K-median algorithm, with probability at least $1-\beta$, $S^{(i)}_\epsilon$ is a $(w, t)$-approximation of $OPT_i$--the optimal K-median with a center fixed at position $z_i$, in which:
  \begin{align*}
    w &= 8(1+\alpha)\\
    t &= 8\bigO_{p,\alpha}\left((k/\beta)^{O_{p,\alpha}(1)}.polylog(n/\beta)/\epsilon\right).
  \end{align*}
\end{corollary}
% \textbf{Application to $\epsilon$-DP K-median clustering with explanation.} \Arielnote{Should be deleted or waiting for more writing?}

%%% Local Variables:
%%% mode: latex
%%% TeX-master: "main"
%%% End:

%\section{K-means with a fixed center}
\subsection{\textsc{NonPrivateApproxFC} for $k$-means}
\label{sec:kmeans}

In this section, we present our \textsc{NonPrivateApproxFC} algorithm for k-means with a fixed center. Based on ~\cite{kanungo2002local}, we achieve a 25-approximation. We will analyze this approximation factor in detail below.
We adapt the work by~\cite{kanungo2002local} by adding a fixed center constraint to the single-swap heuristic algorithm. As in their result, we need to assume that we are given a discrete set of candidate centers $C$ from which we choose $k$ centers. The optimality is defined in the space of all feasible solutions in $C$, i.e., over all subsets of size $k$ of $C$. We then present how to remove this assumption, with the cost of a small constant additive factor.

% \anil{change notation: $C$ is being used for a fixed center in the previous subsection} replaced it with z

\begin{definition}
    \label{def:approx-candidate-center}
    Let $O = (O_1, O_2, \ldots, O_k)$ be the optimal clustering with $O_1$ be the cluster with the fixed center $\tred{z}$.
    A set $C\subset\R^d$ is a $\gamma$-approximate candidate center set if there exists $\tred{z}\in \{c_1, c_2, \ldots, c_k\} \subseteq C$, such that: $cost(c_1, c_2, \ldots, c_k) \leq (1+\gamma)cost(O).$
\end{definition}

\noindent Given \( u, v \in \mathbb{R}^d \), let \( \Delta(u, v) \) denote the squared Euclidean distance between \( u \) and \( v \): \( \Delta(u, v) = \text{dist}^2(u, v) \). For a set \( S \subset \mathbb{R}^d \), the total squared distance between all points in \( S \) and a point \( v \) is given by \( \Delta(S, v) = \sum_{u \in S} \Delta(u, v) \). Similarly, for a set \( P \subset \mathbb{R}^d \), \( \Delta_P(S) \) represents the total squared distance between each point \( q \in P \) and its closest point \( s_q \in S \). Here, \( q \) refers to an individual data point in set \( P \), and \( s_q \) is its nearest neighbor in \( S \). When the context is clear, we drop \( P \) for simplicity. This notation captures the essential relationships between points and their nearest centroids.

Let $\tred{z}$ be the fixed center that must be in the output. Let $C$ be the set of candidate centers, that $\tred{z} \in C$. We define \textbf{stability} in the context of $k$-means with a fixed center $\tred{z}$ as follows. We note that it differs from the definition of~\cite{kanungo2002local} such that we never swap out the fixed center $\tred{z}$:

\begin{definition}
  A set $S$ of k centers that contains the fixed center $\tred{z}$ is called $1$-stable if:
    $\Delta\big(S \setminus \{s\} \cup \{o\}\big) \geq \Delta(S),$
  for all $s\in S\setminus\{\tred{z}\}$, $o \in O \setminus\{\tred{z}\}$.
\end{definition}

\noindent\textbf{Algorithm.} We initialize $S^{(0)}$ as a set of $k$ centers form $C$ that $\tred{z} \in S^{(0)}$. For each set $S^{(i)}$, we perform the swapping iteration:

\begin{itemize}
\item Select one center $s\in S^{(i)} \setminus \tred{z}$
\item Select one replaced center $s' \in C \setminus S^{(i)}$
\item Let $S' = S^{(i)} \setminus s \cup s'$
\item If $S'$ reduces the distortion, $S^{(i+1)} = S'$. Else, $S^{(i+1)} = S^{(i)}$
\end{itemize}

We repeat the swapping iteration until $S = S^{(m)}$, i.e., after $m$ iterations, is a $1$-stable.
%\Arielnote{Accordingly to lemma 7(not [10])} set.
Theorem~\ref{theorem:kmean-utility} states the utility of an arbitrary $1$-stable set, which is also the utility of our algorithm since it always outputs an $1$-stable set. We note that if $C$ is created with some errors $\gamma$ to the actual optimal centroids, the utility bound of our algorithm is increased by the factor $\Theta(\gamma)$, i.e., ours is a $(25+\Theta(\gamma))$-approximation to the actual optimal centroids.

\begin{theorem} {\bf Approximation factor of fixed centroid k-mean.}
  \label{theorem:kmean-utility}
  If $S$ is an $1$-stable $k$-element set of centers, $\Delta(S) \leq 25\Delta(O)$. Furthermore, if $C$ is a $\frac{\gamma}{25}$-approximate candidate center set, $S$ is a $(25+\gamma)$-approximate of the actual optimal centroids in the Euclidean space.
\end{theorem}
Having demonstrated that our modification of~\cite{kanungo2002local} to execute fixed-centroid k-means instead of standard k-means yields a 25-factor approximation of the optimal solution, we can now proceed to prove that our private explanation closely approximates the optimal solution for the fixed-centroid scenario.
\begin{corollary}
  \label{cor:private-kmeans-utility}
  Running \algo~with \textsc{NonPrivateApproxFC} be the above $k$-means algorithm, with probability at least $1-\beta$, $S^{(i)}_\epsilon$ is a $(w, t)$-approximation of $OPT_i$--the optimal $k$-means with a center fixed at position $x_i$, in which:
  \begin{align*}
    w &= (25+\gamma)(1+\alpha)\\
    t &= (25+\gamma)\bigO_{p,\alpha}\left((k/\beta)^{O_{p,\alpha}(1)}.polylog(n/\beta)/\epsilon\right).
  \end{align*}
\end{corollary}

\noindent
 With the utility bounds for k-means and k-median under the fixed-centroid constraint proven, it is clear that altering the original algorithms preserves the same utility bounds as their non-fixed counterparts. This ensures that accommodating fixed centroids does not compromise clustering quality. Notably, these bounds refer to clustering utility, not explanation bounds. By showing that fixed-k-means and fixed-k-median perform as effectively as standard versions, users can trust the quality of the explanations. Without these guarantees, users might question the validity of centroid placements. Our results ensure the explanations are based on clustering solutions that are as robust and reliable as the original algorithms.

%%% Local Variables:
%%% mode: latex
%%% TeX-master: "main"
%%% End:

% \begin{figure*}[htbp]
%     \centering
%     \includegraphics[width=\textwidth,keepaspectratio]{new-exp/dimensions.png}
%         \label{fig:dimensions}
% \end{figure*}

\begin{figure*}[htbp]
    \centering
    \includegraphics[width=.9\textwidth,keepaspectratio]{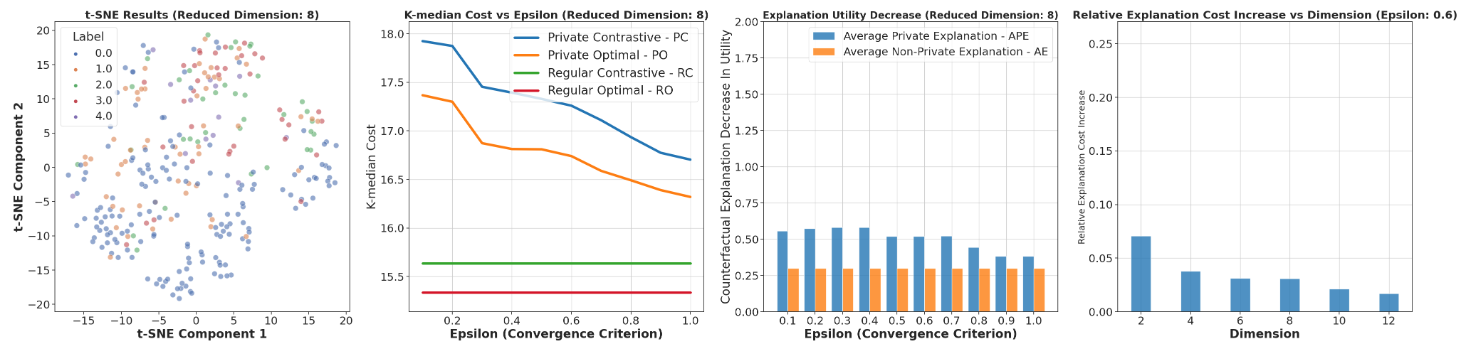}
    \caption{
        A visualization of our dataset (Heart Disease dataset from the UCI ML Repository), projected into an 8-dimensional space. (a) t-SNE of our data. (b) Comparison of k-median clustering with fixed and non-fixed centroids, both private and non-private. (c) Bar graph showing contrastive explanation differences for differential private and non-private k-median with a fixed centroid. (d) We fix the privacy budget of 0.6 while demonstrating the contrastive explanation across various dimensions.
        }
    \label{fig:United}
\end{figure*}

\section{Experiments}
%This phase is crucial for assessing the practical applicability of our method 
%in real-life scenarios.
Our study examines how the privacy budget $\epsilon$ affects the trade-off between privacy and accuracy, focusing on the quality of differentially private explanations. We use four key metrics: Private Optimal (PO, $S_\epsilon$), Private Contrastive (PC, $S^{(i)}_\epsilon$), Regular Optimal (RO, $OPT$), and Regular Contrastive (RC, $OPT_i$), to compare clustering costs with and without fixed centroids in both private and non-private algorithms.
To assess explanation quality, we define two derived metrics: Average Private Explanation (APE, PC - PO, cost($S^{(i)}_\epsilon$) - cost($S_\epsilon$)) and Average Explanation (AE, RC - RO, cost($S^{(i)}$) - cost($S$)). APE measures utility loss in private clustering as an explanatory output, while AE provides a non-private baseline. These metrics help us evaluate the explanatory power of our approach.
By analyzing these metrics across different $\epsilon$ values, we explore the balance between privacy and utility, highlighting the trade-offs in our differentially private clustering and explanation framework.

\textbf{Datasets}
Our research utilizes a diverse set of datasets to demonstrate the versatility and effectiveness of our approach, as summarized in Table \ref{tab:datasets}. We employed the Heart Disease dataset featuring 13 dimensions, and the Breast Cancer dataset with 30 features, including both numeric and categorical fields. Both datasets were taken from the UCI Machine Learning Repository Those higher-dimensional datasets were crucial in validating our theoretical framework. Additionally, we used two-dimensional activity-based population datasets from Charlottesville City and Albemarle County, Virginia, previously employed in mobile vaccine clinic deployment studies \cite{mehrab2022data}. To complement these real-world datasets, we also generated a synthetic two-dimensional dataset. By testing our method on both high-dimensional and two-dimensional data, as well as on real and synthetic datasets, we showcase its robustness across different data complexities and origins.
\begin{table}[b]
\centering
\small
\begin{tabular}{lccc}
\hline
\textbf{Dataset} & \textbf{Dim} & \textbf{Size} & \textbf{Source} \\
\hline
Heart Disease & 13 & 303 & UCI MLR \\
Breast Cancer Wisconsin & 31 & 569 & UCI MLR \\
Charlottesville & 2 & 33K & {\cite{mehrab2022data}} \\
Albemarle & 2 & 74K & {\cite{mehrab2022data}} \\
Synthetic dataset & 2 & 1k & Generated \\
\hline
\end{tabular}
\caption{Datasets used in our research}
\label{tab:datasets}
\end{table}
\begin{figure*}[h]
    \centering
    \includegraphics[width=0.70\textwidth,keepaspectratio]{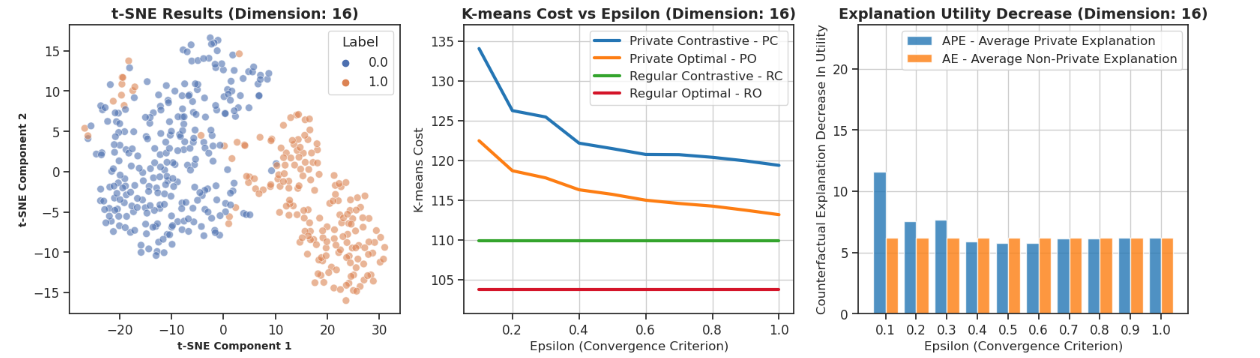}
        \caption{
         A visualization of another dataset (Breast Cancer dataset (30 features) from the UCI ML Repository), projected into a 16-dimensional space. (a) t-SNE of our data. (b) Comparison of k-means clustering with fixed and non-fixed centroids, both private and non-private. (c) Bar graph showing contrastive explanation differences for differential private and non-private k-means with a fixed centroid.}
        \label{fig:BreastCancer}
\end{figure*}

\noindent \textbf{Data Preprocessing:} We normalized all datasets to fit within a unit ball to ensure consistency with prior work and standardize our analysis framework. While this normalization alters the absolute scale, it preserves the relative relationships between data points, which is crucial for clustering. The entire preprocessed dataset was used for analysis, as there is no ground truth labeling for a traditional train-test split in this unsupervised task.

\textbf{Dimensionality} Our study explored both 2D and higher-dimensional datasets. A crucial aspect of our methodology, \textsc{DimReduction}, employs Principal Component Analysis (PCA) for initial dimensionality reduction. This process normalizes the data and creates lower-dimensional representations. We performed extensive experiments, reducing high-dimensional datasets to various lower dimensions, including 2D, with additional low-dimension experiments.
Remarkably, our results remained consistent across different reduced dimensionalities. Even when reducing data from 13 dimensions to 2, we observed similar trends and results as with other dimensional reductions, despite significant information loss. This consistency underscores our method's robustness across varying dimensions.
By addressing high-dimensional data challenges through PCA reduction, we ensure our technique's applicability and efficiency across diverse dataset complexities, maintaining result integrity regardless of original data dimensionality.

\textbf{Running Time Analysis:} The computational complexity of our algorithm varies by clustering method. For k-means, we use the linear-time algorithm from ~\cite{kanungo2002local}, while the k-median approach relies on polynomial-time Linear Programming (LP). We have optimized performance with GPU parallelization, reducing execution times from minutes to seconds for both differentially private coresets and clustering tasks. Our method is \textbf{data-agnostic}, handling any data distribution efficiently, independent of sparsity. For reproducibility, we provide our code, experimental details, and pre-processed datasets in a public repository.
\subsection{Experimental results}
\noindent
Figure~\ref{fig:United} presents four key visualizations of our differentially private clustering and explanation framework across various dimensions.
The t-SNE plot (leftmost) shows the 2D representation of our dataset, revealing potential clusters and patterns. The second plot illustrates K-median cost versus $\epsilon$ for our four metrics (PC, PO, RC, RO), demonstrating the privacy-accuracy trade-off and the consistency of our contrastive explanations.

In plots (b) and (c) of Figure 1, the x-axis represents the privacy budget \(\epsilon\), which we tested over the range \([0, 1]\) in intervals of 0.05. This granularity allows for a detailed analysis of the privacy-utility trade-off. Smaller values of \(\epsilon\) enforce stronger privacy guarantees, as reflected in higher clustering costs for PC and PO. Conversely, as \(\epsilon\) increases, these costs gradually decrease, highlighting the improved utility that comes with relaxed privacy constraints. Importantly, the observed trends demonstrate that the framework consistently balances privacy and utility, even at stricter privacy levels. Additionally, the stability of the non-private metrics (RC and RO) across the \(\epsilon\) range provides a robust baseline for evaluating the performance of our private clustering and explanation methods.

As expected, both PC and PO costs decrease as $\epsilon$ increases, demonstrating the trade-off between privacy and accuracy. The non-private metrics (RC and RO) remain constant across $\epsilon$ values, serving as baselines for comparison. Notably, the gap between PC and PO remains relatively consistent, indicating that our contrastive explanations maintain their relative quality at different levels of privacy.
The third plot illustrates the Explanation Utility for both private (Average Private Explanation) and non-private (Average Non-Private Explanation) scenarios across various $\epsilon$ values. This graph quantifies the difference in clustering cost between the optimal solution and the solution with a fixed centroid, representing our contrastive explanations.
Notably, we observe that the Average Private Explanation remains relatively stable across different $\epsilon$ values. This stability is crucial as it indicates that the quality of our contrastive explanations in the private setting is consistent, regardless of the privacy budget.
 The consistent performance across different $\epsilon$ values underscores the robustness of our method, providing reliable explanations even under strict privacy constraints.
 The rightmost plot demonstrates the difference between PC and PO across dimensions for a fixed $\epsilon$, illustrating our method's scalability with dimensionality.
 
Figure~\ref{fig:BreastCancer} follows a similar format but uses another high-dimensional dataset. This dataset was reduced from 30 dimensions to 16 to test the robustness of our approach on different datasets and higher dimensions. Unlike Figure 1, this figure presents results obtained using the k-means algorithm. Furthermore, we extended our experiments to include other reduced dimensions for both k-means and k-median, with the detailed results provided in the Appendix. These additional experiments further validate the adaptability and robustness of our framework across different clustering methods and dimensionality settings.

\section{Conclusions}
Our work explores the design of private explanations for clustering, particularly focusing on the k-median and k-means objectives for Euclidean datasets. We formalize this as the PRIVEC problem, where each agent receives a contrastive explanation corresponding to the loss in utility they experience when a cluster centroid is placed at a strategic position chosen by the agent.
Our algorithm provides explanations to each user while maintaining the same approximation factor as private clustering, within a predefined privacy budget. The related work in this domain has shown the development of algorithms for contrastive explanations, but our contribution stands out by integrating differential privacy guarantees. 

Our experiments demonstrate the resilience of our approach. Despite the added layer of providing differentially private explanations on top of differentially private clustering, the quality of our explanations remains uncompromised.
The extended experiments on all our datasets further validate our approach's efficacy. The balance between privacy and utility, the robustness of contrastive explanations, and the negligible impact of $\epsilon$ on explainability were consistent across datasets. These findings underscore the potential of our method for diverse real-world applications.

Our approach is not restricted to k-means and k-median but can be applied to other clustering algorithms as well. The methodology leverages fundamental principles common to many clustering techniques, such as centroids and utility functions. As long as a clustering algorithm defines centroids and evaluates clustering quality using these metrics, our approach can be adapted to provide privacy-preserving contrastive explanations. This adaptability makes it suitable for extending to other paradigms, such as density-based or hierarchical clustering, extending its applicability to various datasets and contexts.
  
%%% Local Variables:
%%% mode: latex
%%% TeX-master: "main"
%%% End:

\textbf{Acknowledgment}
This research is partially supported by the Israel
Ministry of Innovation, Science \& Technology grant
1001818511, NSF grants CCF-1918656, CNS-2317193, IIS-2331315, and CDC MIND U01CK000589.

\clearpage
\bibliographystyle{plainnat}
\bibliography{refs.bib}

\begin{thebibliography}{42}
\providecommand{\natexlab}[1]{#1}
\providecommand{\url}[1]{\texttt{#1}}
\expandafter\ifx\csname urlstyle\endcsname\relax
  \providecommand{\doi}[1]{doi: #1}\else
  \providecommand{\doi}{doi: \begingroup \urlstyle{rm}\Url}\fi

\bibitem[Balcan et~al.(2017)Balcan, Dick, Liang, Mou, and
  Zhang]{DBLP:conf/icml/BalcanDLMZ17}
Maria{-}Florina Balcan, Travis Dick, Yingyu Liang, Wenlong Mou, and Hongyang
  Zhang.
\newblock Differentially private clustering in high-dimensional euclidean
  spaces.
\newblock In Doina Precup and Yee~Whye Teh, editors, \emph{Proceedings of the
  34th International Conference on Machine Learning, {ICML} 2017, Sydney, NSW,
  Australia, 6-11 August 2017}, volume~70 of \emph{Proceedings of Machine
  Learning Research}, pages 322--331. {PMLR}, 2017.
\newblock URL \url{http://proceedings.mlr.press/v70/balcan17a.html}.

\bibitem[Bobek et~al.(2022)Bobek, Kuk, Szel{\k{a}}{\.z}ek, and
  Nalepa]{bobek2022enhancing}
Szymon Bobek, Michal Kuk, Maciej Szel{k{a}}{\.z}ek, and Grzegorz~J Nalepa.
\newblock Enhancing cluster analysis with explainable ai and multidimensional
  cluster prototypes.
\newblock \emph{IEEE Access}, 10:\penalty0 101556--101574, 2022.

\bibitem[Boggess et~al.(2022)Boggess, Kraus, and Feng]{boggess2022toward}
K~Boggess, S~Kraus, and L~Feng.
\newblock Toward policy explanations for multi-agent reinforcement learning.
\newblock In \emph{International Joint Conference on Artificial Intelligence
  (IJCAI)}, 2022.

\bibitem[Boggess et~al.(2023)Boggess, Kraus, and Feng]{boggess2023explainable}
Kayla Boggess, Sarit Kraus, and Lu~Feng.
\newblock Explainable multi-agent reinforcement learning for temporal queries.
\newblock In \emph{Proceedings of the Thirty-Second International Joint
  Conference on Artificial Intelligence (IJCAI)}, 2023.

\bibitem[Charikar et~al.(1999)Charikar, Guha, Tardos, and
  Shmoys]{charikar1999constant}
Moses Charikar, Sudipto Guha, {\'E}va Tardos, and David~B Shmoys.
\newblock A constant-factor approximation algorithm for the k-median problem.
\newblock In \emph{Proceedings of the thirty-first annual ACM symposium on
  Theory of computing}, pages 1--10, 1999.

\bibitem[Chen et~al.(2023)Chen, Cohen-Addad, d’Orsi, Epasto, Imola, Steurer,
  and Tiegel]{chen2023private}
Hongjie Chen, Vincent Cohen-Addad, Tommaso d’Orsi, Alessandro Epasto, Jacob
  Imola, David Steurer, and Stefan Tiegel.
\newblock Private estimation algorithms for stochastic block models and mixture
  models.
\newblock \emph{Advances in Neural Information Processing Systems},
  36:\penalty0 68134--68183, 2023.

\bibitem[Dasgupta and Gupta(2003)]{dasgupta2003elementary}
Sanjoy Dasgupta and Anupam Gupta.
\newblock An elementary proof of a theorem of johnson and lindenstrauss.
\newblock \emph{Random Structures \& Algorithms}, 22\penalty0 (1):\penalty0
  60--65, 2003.

\bibitem[Dwork and Roth(2014)]{dwork_and_roth}
Cynthia Dwork and Aaron Roth.
\newblock The algorithmic foundations of differential privacy.
\newblock \emph{Found. Trends Theor. Comput. Sci.}, 9\penalty0
  (3–4):\penalty0 211–407, aug 2014.
\newblock ISSN 1551-305X.
\newblock \doi{10.1561/0400000042}.
\newblock URL \url{https://doi.org/10.1561/0400000042}.

\bibitem[Feldman et~al.(2017)Feldman, Xiang, Zhu, and Rus]{feldman2017coresets}
Dan Feldman, Chongyuan Xiang, Ruihao Zhu, and Daniela Rus.
\newblock Coresets for differentially private k-means clustering and
  applications to privacy in mobile sensor networks.
\newblock In \emph{Proceedings of the 16th ACM/IEEE International Conference on
  Information Processing in Sensor Networks}, pages 3--15, 2017.

\bibitem[Finkelstein et~al.(2022)Finkelstein, Liu, Kolumbus, Parkes,
  Rosenschein, Keren, et~al.]{finkelstein2022explainable}
Mira Finkelstein, Lucy Liu, Yoav Kolumbus, David~C Parkes, Jeffrey~S
  Rosenschein, Sarah Keren, et~al.
\newblock Explainable reinforcement learning via model transforms.
\newblock \emph{Advances in Neural Information Processing Systems},
  35:\penalty0 34039--34051, 2022.

\bibitem[Georgara et~al.(2022)Georgara, Rodr{\'\i}guez-Aguilar, and
  Sierra]{georgara2022privacy}
Athina Georgara, Juan~Antonio Rodr{\'\i}guez-Aguilar, and Carles Sierra.
\newblock Privacy-aware explanations for team formation.
\newblock In \emph{International Conference on Principles and Practice of
  Multi-Agent Systems}, pages 543--552. Springer, 2022.

\bibitem[Ghadiri et~al.(2021)Ghadiri, Samadi, and Vempala]{ghadiri2021socially}
Mehrdad Ghadiri, Samira Samadi, and Santosh Vempala.
\newblock Socially fair k-means clustering.
\newblock In \emph{Proceedings of the 2021 ACM Conference on Fairness,
  Accountability, and Transparency}, pages 438--448, 2021.

\bibitem[Ghazi et~al.(2020)Ghazi, Kumar, and Manurangsi]{ghazi:neurips20}
Badih Ghazi, Ravi Kumar, and Pasin Manurangsi.
\newblock Differentially private clustering: Tight approximation ratios.
\newblock In \emph{Proceedings of the 34th International Conference on Neural
  Information Processing Systems}, NIPS'20, Red Hook, NY, USA, 2020. Curran
  Associates Inc.
\newblock ISBN 9781713829546.

\bibitem[Goethals et~al.(2022)Goethals, S{\"o}rensen, and
  Martens]{goethals2022privacy}
Sofie Goethals, Kenneth S{\"o}rensen, and David Martens.
\newblock The privacy issue of counterfactual explanations: explanation linkage
  attacks.
\newblock \emph{arXiv preprint arXiv:2210.12051}, 2022.

\bibitem[Gupta et~al.(2010)Gupta, Ligett, McSherry, Roth, and
  Talwar]{gupta2009differentially}
Anupam Gupta, Katrina Ligett, Frank McSherry, Aaron Roth, and Kunal Talwar.
\newblock Differentially private combinatorial optimization.
\newblock In \emph{Proceedings of the Twenty-First Annual {ACM-SIAM} Symposium
  on Discrete Algorithms, {SODA} 2010, Austin, Texas, USA, January 17-19,
  2010}, pages 1106--1125. {SIAM}, 2010.
\newblock \doi{10.1137/1.9781611973075.90}.
\newblock URL \url{https://doi.org/10.1137/1.9781611973075.90}.

\bibitem[Huang and Liu(2018)]{huang2018optimal}
Zhiyi Huang and Jinyan Liu.
\newblock Optimal differentially private algorithms for k-means clustering.
\newblock In \emph{Proceedings of the 37th ACM SIGMOD-SIGACT-SIGAI Symposium on
  Principles of Database Systems}, pages 395--408, 2018.

\bibitem[Ji et~al.(2019)Ji, Luo, Guo, Ji, Liao, and Li]{ji2019differentially}
Tianxi Ji, Changqing Luo, Yifan Guo, Jinlong Ji, Weixian Liao, and Pan Li.
\newblock Differentially private community detection in attributed social
  networks.
\newblock In \emph{Asian Conference on Machine Learning}, pages 16--31. PMLR,
  2019.

\bibitem[Johnson and Lindenstrauss(1984)]{johnson1984extensions}
William~B Johnson and Joram Lindenstrauss.
\newblock Extensions of lipschitz mappings into a hilbert space.
\newblock In \emph{Conference on Modern Analysis and Probability}, volume~26,
  pages 189--206. American Mathematical Society, 1984.

\bibitem[Kanungo et~al.(2002)Kanungo, Mount, Netanyahu, Piatko, Silverman, and
  Wu]{kanungo2002local}
Tapas Kanungo, David~M Mount, Nathan~S Netanyahu, Christine~D Piatko, Ruth
  Silverman, and Angela~Y Wu.
\newblock A local search approximation algorithm for k-means clustering.
\newblock In \emph{Proceedings of the eighteenth annual symposium on
  Computational geometry}, pages 10--18, 2002.

\bibitem[Madumal et~al.(2020)Madumal, Miller, Sonenberg, and
  Vetere]{madumal2020explainable}
Prashan Madumal, Tim Miller, Liz Sonenberg, and Frank Vetere.
\newblock Explainable reinforcement learning through a causal lens.
\newblock In \emph{Proceedings of the AAAI conference on artificial
  intelligence}, volume~34, pages 2493--2500, 2020.

\bibitem[Makarychev et~al.(2019)Makarychev, Makarychev, and
  Razenshteyn]{makarychev2019performance}
Konstantin Makarychev, Yury Makarychev, and Ilya Razenshteyn.
\newblock Performance of johnson-lindenstrauss transform for k-means and
  k-medians clustering.
\newblock In \emph{Proceedings of the 51st Annual ACM SIGACT Symposium on
  Theory of Computing}, pages 1027--1038, 2019.

\bibitem[Matou{\v{s}}ek(2000)]{matouvsek2000approximate}
Ji{\v{r}}{\'\i} Matou{\v{s}}ek.
\newblock On approximate geometric k-clustering.
\newblock \emph{Discrete \& Computational Geometry}, 24\penalty0 (1):\penalty0
  61--84, 2000.

\bibitem[Mehrab et~al.(2022)Mehrab, Wilson, Chang, Harrison, Lewis, Telionis,
  Crow, Kim, Spillmann, Peters, et~al.]{mehrab2022data}
Zakaria Mehrab, Mandy~L Wilson, Serina Chang, Galen Harrison, Bryan Lewis, Alex
  Telionis, Justin Crow, Dennis Kim, Scott Spillmann, Kate Peters, et~al.
\newblock Data-driven real-time strategic placement of mobile vaccine
  distribution sites.
\newblock In \emph{Proceedings of the AAAI Conference on Artificial
  Intelligence}, volume~36, pages 12573--12579, 2022.

\bibitem[Miller(2019)]{miller2019explanation}
Tim Miller.
\newblock Explanation in artificial intelligence: Insights from the social
  sciences.
\newblock \emph{Artificial intelligence}, 267:\penalty0 1--38, 2019.

\bibitem[Moshkovitz et~al.(2020)Moshkovitz, Dasgupta, Rashtchian, and
  Frost]{moshkovitz2020explainable}
Michal Moshkovitz, Sanjoy Dasgupta, Cyrus Rashtchian, and Nave Frost.
\newblock Explainable k-means and k-medians clustering.
\newblock In \emph{International conference on machine learning}, pages
  7055--7065. PMLR, 2020.

\bibitem[Newling and Fleuret(2016)]{newling2016fast}
James Newling and Fran{\c{c}}ois Fleuret.
\newblock Fast k-means with accurate bounds.
\newblock In \emph{International Conference on Machine Learning}, pages
  936--944. PMLR, 2016.

\bibitem[Nguyen and Vullikanti(2024)]{pmlr-v235-nguyen24j}
Dung Nguyen and Anil~Kumar Vullikanti.
\newblock Differentially private exact recovery for stochastic block models.
\newblock In Ruslan Salakhutdinov, Zico Kolter, Katherine Heller, Adrian
  Weller, Nuria Oliver, Jonathan Scarlett, and Felix Berkenkamp, editors,
  \emph{Proceedings of the 41st International Conference on Machine Learning},
  volume 235 of \emph{Proceedings of Machine Learning Research}, pages
  37798--37839. PMLR, 21--27 Jul 2024.
\newblock URL \url{https://proceedings.mlr.press/v235/nguyen24j.html}.

\bibitem[Nguyen et~al.(2024)Nguyen, Vetzler, Kraus, and
  Vullikanti]{nguyen2024contrastiveexplainableclusteringdifferential}
Dung Nguyen, Ariel Vetzler, Sarit Kraus, and Anil Vullikanti.
\newblock Contrastive explainable clustering with differential privacy, 2024.
\newblock URL \url{https://arxiv.org/abs/2406.04610}.

\bibitem[Nguyen et~al.(2023)Nguyen, Lai, Phan, and Thai]{nguyen2023xrand}
Truc Nguyen, Phung Lai, Hai Phan, and My~T Thai.
\newblock Xrand: Differentially private defense against explanation-guided
  attacks.
\newblock In \emph{Proceedings of the AAAI Conference on Artificial
  Intelligence}, volume~37, pages 11873--11881, 2023.

\bibitem[Nissim and Stemmer(2018)]{nissim2018clustering}
Kobbi Nissim and Uri Stemmer.
\newblock Clustering algorithms for the centralized and local models.
\newblock In \emph{Algorithmic Learning Theory}, pages 619--653. PMLR, 2018.

\bibitem[Ofek and Somech(2024)]{ofek2024explaining}
Sariel Ofek and Amit Somech.
\newblock Explaining black-box clustering pipelines with cluster-explorer.
\newblock \emph{arXiv preprint arXiv:2412.20446}, 2024.

\bibitem[Patel et~al.(2022)Patel, Shokri, and Zick]{patel2022model}
Neel Patel, Reza Shokri, and Yair Zick.
\newblock Model explanations with differential privacy.
\newblock In \emph{Proceedings of the 2022 ACM Conference on Fairness,
  Accountability, and Transparency}, pages 1895--1904, 2022.

\bibitem[Pozanco et~al.(2022)Pozanco, Mosca, Zehtabi, Magazzeni, and
  Kraus]{pozanco2022explaining}
Alberto Pozanco, Francesca Mosca, Parisa Zehtabi, Daniele Magazzeni, and Sarit
  Kraus.
\newblock Explaining preference-driven schedules: the expres framework.
\newblock In \emph{Proceedings of the International Conference on Automated
  Planning and Scheduling}, volume~32, pages 710--718, 2022.

\bibitem[Reddy(2018)]{reddy2018data}
Chandan~K Reddy.
\newblock \emph{Data clustering: algorithms and applications}.
\newblock Chapman and Hall/CRC, 2018.

\bibitem[Saifullah et~al.(2022)Saifullah, Mercier, Lucieri, Dengel, and
  Ahmed]{saifullah2022privacy}
Saifullah Saifullah, Dominique Mercier, Adriano Lucieri, Andreas Dengel, and
  Sheraz Ahmed.
\newblock Privacy meets explainability: A comprehensive impact benchmark.
\newblock \emph{arXiv preprint arXiv:2211.04110}, 2022.

\bibitem[Schleibaum et~al.(2024)Schleibaum, Feng, Kraus, and
  M{\"u}ller]{schleibaum2024adesse}
S{\"o}ren Schleibaum, Lu~Feng, Sarit Kraus, and J{\"o}rg~P M{\"u}ller.
\newblock Adesse: Advice explanations in complex repeated decision-making
  environments.
\newblock \emph{arXiv preprint arXiv:2405.20705}, 2024.

\bibitem[Sreedharan et~al.(2020)Sreedharan, Soni, Verma, Srivastava, and
  Kambhampati]{sreedharan2020bridging}
Sarath Sreedharan, Utkarsh Soni, Mudit Verma, Siddharth Srivastava, and
  Subbarao Kambhampati.
\newblock Bridging the gap: Providing post-hoc symbolic explanations for
  sequential decision-making problems with inscrutable representations.
\newblock \emph{arXiv preprint arXiv:2002.01080}, 2020.

\bibitem[Sreedharan et~al.(2021)Sreedharan, Srivastava, and
  Kambhampati]{sreedharan2021using}
Sarath Sreedharan, Siddharth Srivastava, and Subbarao Kambhampati.
\newblock Using state abstractions to compute personalized contrastive
  explanations for ai agent behavior.
\newblock \emph{Artificial Intelligence}, 301:\penalty0 103570, 2021.

\bibitem[Stemmer(2020)]{stemmer:soda20}
Uri Stemmer.
\newblock Locally private \emph{k}-means clustering.
\newblock In Shuchi Chawla, editor, \emph{Proceedings of the 2020 {ACM-SIAM}
  Symposium on Discrete Algorithms, {SODA} 2020, Salt Lake City, UT, USA,
  January 5-8, 2020}, pages 548--559. {SIAM}, 2020.
\newblock \doi{10.1137/1.9781611975994.33}.
\newblock URL \url{https://doi.org/10.1137/1.9781611975994.33}.

\bibitem[Stemmer and Kaplan(2018)]{stemmer2018differentially}
Uri Stemmer and Haim Kaplan.
\newblock Differentially private k-means with constant multiplicative error.
\newblock \emph{Advances in Neural Information Processing Systems}, 31, 2018.

\bibitem[van~der Waa et~al.(2018)van~der Waa, van Diggelen, Bosch, and
  Neerincx]{van2018contrastive}
Jasper van~der Waa, Jurriaan van Diggelen, Karel van~den Bosch, and Mark
  Neerincx.
\newblock Contrastive explanations for reinforcement learning in terms of
  expected consequences.
\newblock \emph{arXiv preprint arXiv:1807.08706}, 2018.

\bibitem[Zehtabi et~al.(2024)Zehtabi, Pozanco, Bolch, Borrajo, and
  Kraus]{zehtabi2024contrastive}
Parisa Zehtabi, Alberto Pozanco, Ayala Bolch, Daniel Borrajo, and Sarit Kraus.
\newblock Contrastive explanations of centralized multi-agent optimization
  solutions.
\newblock In \emph{Proceedings of the International Conference on Automated
  Planning and Scheduling}, volume~34, pages 671--679, 2024.

\end{thebibliography}

\appendix

\onecolumn

\begin{table*}[!ht]
    \centering
    \begin{tabular}{| c |c | c |}
      \hline
      Notations & Definitions &Note \\
      \hline
      $X \sim X'$& Two datasets $X$ and $X'$ differ by at most $1$ element & \\
      %\textsc{BaselinePE} (Algorithm~\ref{alg:dp-explain-baseline}) & The baseline algorithm using black-box DP clustering and& Page~\pageref{alg:dp-explain-baseline}\\
      % & applying composition to answer multiple requests&\\
      $\alpha$ & Approximation parameter for the utility of clustering and explanations& \\
      $\beta$ & Failure probability of the utility guarantees of clustering and explanations& \\
      $\zeta$ & Parameter to control the utility of the private coreset: $\zeta = 0.01\left(\frac{\alpha}{10\lambda_{p,\alpha/2}}\right)^{1/p}$&  \\
      $\lambda_{p,\alpha/2}$ & Definition~\ref{def:lambda}&  \\
      $\bigO_{p,\alpha}$ &Big O notation that explicit ignore factors of $p$ and $\alpha$&\\
      $w, t$ &Approximation factors of the utility of our private explanations& Def.~\ref{def:wt-approx}\\
      $w''$ &Approximation factors of the non-private clustering algorithm when one centroid is fixed&\\
      $OPT$ &Cost of the optimal clustering& \\
      $OPT_i$ &Cost the the optimal clustering when one centroid is fixed at requested location of agent $i$ &\\
      $d, d'$ &The \# of dimensions of the original and projected spaces& \\
      $S_{\epsilon}$ &Private clustering solution with privacy budget $\epsilon$&\\
      $S^{(i)}_{\epsilon}$ &Private solution computed by the clustering algorithm while &\\
      &Fixing one centroid to the requested position of agent $i$&\\
      \textsc{NonPrivateApprox} &Any (not necessarily private) clustering algorithm with approximation factor $w'\leq w''$&\\
      \textsc{NonPrivateApproxFC} &Any (not necessarily private) clustering algorithm with one centroid fixed as request,&\\
      &with approximation factor $w''$&\\
      $X$ &Input dataset in the original space&\\
      $X'$ &Projected dataset in the $d'$-dimensional space& in Theorem~\ref{theorem:explanation-utility} \\
      $Y$ &The private coreset&\\
      $\mathcal{S}$ & Projection from $\R^d$ to $\mathcal{S}$&\\
      \hline
    \end{tabular}
    \caption{Summary of repeatedly used notations and their definitions}
  \label{table:notation}
\end{table*}

\section{Related work: additional details}
\label{sec:related-appendix}

Our work considers differential privacy for explainable AI in general (XAI) and Multi-agent explanations (XMASE) in particular, focusing on post-hoc contrastive explanations for clustering. 

Extensive experiments presented in \cite{saifullah2022privacy} demonstrate non-negligible changes in
explanations of black-box ML models through the introduction of privacy. The findings in~\cite{patel2022model} corroborate these observations regarding explanations for black-box feature-based models. These explanations involve creating local approximations of the model's behavior around specific points of interest, potentially utilizing sensitive data.
In order to safeguard the privacy of the data used during the local approximation process of an eXplainable Artificial Intelligence (XAI) module, the researchers have devised an innovative adaptive differentially private algorithm. This algorithm is designed to determine the minimum privacy budget required to generate accurate explanations effectively. The study undertakes a comprehensive evaluation, employing both empirical and analytical methods, to assess how the introduction of randomness inherent in differential privacy algorithms impacts the faithfulness of the model explanations.

\cite{nguyen2023xrand} considers feature-based explanations (e.g., SHAP) that can expose
the top important features that a black-box model focuses on. To prevent such expose 
they introduced a new concept of achieving local differential
privacy (LDP) in the explanations, and from that, they established
a defense, called XRAND, against such attacks. They showed that
their mechanism restricts the information that the adversary
can learn about the top important features while maintaining
the faithfulness of the explanations. 

The analysis presented in restatable\cite{goethals2022privacy} considers security concerning contrastive explanations. The authors introduced the concept of the "explanation linkage attack", a potential vulnerability that arises when employing instance-based strategies to derive contrastive explanations. To address this concern, they put forth the notion of k-anonymous contrastive explanations.
Furthermore, the study highlights the intricate balance between transparency, fairness, and privacy when incorporating k-anonymous explanations. As the degree of privacy constraints is heightened, a discernible trade-off comes into play: the quality of explanations and, consequently, transparency are compromised.

Amongst the three types of eXplainable AI mentioned earlier, the maintenance of privacy during explanation generation incurs a certain cost. This cost remains even if an expense was previously borne during the creation of the original model. However, in our proposed methodology for generating contrastive explanations in clustering scenarios, once the cost of upholding differential privacy in the initial solution is paid, no additional expenses are requisite to ensure differential privacy during the explanation generation phase.

Closer to our application is the study that investigates the privacy aspects concerning contrastive explanations in the context of team formation \cite{georgara2022privacy}. In this study, the authors present a comprehensive framework that integrates team formation solutions with their corresponding explanations, while also addressing potential privacy concerns associated with these explanations.
To accomplish this, the authors introduce a privacy breach detector (PBD) that is designed to evaluate whether the provision of an explanation might lead to privacy breaches. The PBD consists of two main components:
(a) A belief updater (BU), calculates the posterior beliefs that a user is likely to form after receiving the explanation.
(b) A privacy checker (PC), examines whether the user's expected posterior beliefs surpass a specified belief threshold, indicating a potential privacy breach.
However, the research is still in its preliminary stages and needs a detailed evaluation of the privacy breach detector.

Our contribution includes the development of comprehensive algorithms for generating contrastive explanations with differential privacy guarantees. We have successfully demonstrated the effectiveness of these algorithms by providing rigorous proof for their privacy guarantees and conducting extensive experiments that showcased their accuracy and utility. In particular, we have shown the validity of our private explanations for clustering based on the $k$-median and $k$-means objectives for Euclidean datasets. Moreover, our algorithms have been proven to have the same accuracy bounds as the best private clustering methods, even though they provide explanations for all users, within a bounded privacy budget. Notably, our experiments in the dedicated experiments section reveal that the epsilon budget has minimal impact on the explainability of our results, further highlighting the robustness of our approach.

There has been a lot of work on private clustering and facility location, starting with ~\cite{gupta2009differentially}, which was followed by a lot of work on other clustering problems in different privacy models, e.g.,~\cite{huang2018optimal,stemmer:soda20,stemmer2018differentially,nissim2018clustering,feldman2017coresets}.
\cite{gupta2009differentially} demonstrated that the additive error bound for points in a metric space involves an $O(\Delta k^2\log{n}/\epsilon)$ term, where $\Delta$ is the space's diameter. Consequently, all subsequent work, including ours, assumes points are restricted to a unit ball.
In addition, there has been extensive work on a closely related problem, in the context of private clustering on graphs or networked data, often mentioned as community detection~\cite{pmlr-v235-nguyen24j,chen2023private,ji2019differentially}. 

% We briefly summarize the work on private clustering and facility location, which is directly relevant to our work.
% One of the first works to study these problems was~\cite{gupta2009differentially}, who showed that the additive error bound for points in a metric space involves an $O(\Delta k^2\log{n}/\epsilon)$ term, where $\Delta$ is the diameter of the space, and this term cannot be avoided.
% As a result, all subsequent work, including ours, assumes the points to be restricted to a unit ball.

% comment out due to not used
% \section{Privacy tools and proofs}
% 
% 
% \composition*
% 
% \begin{proof}
%   For any $O = (o^{(1)}, o^{(2)})$ and any pair of datasets $X, X'$ such that $X\setminus \{Y\cup Z\} \sim X'\setminus \{Y\cup Z\}$, we have:
% 
%   \begin{align}
%     \frac{\Pr[M(X) = O]}{\Pr[M(X') = O]} &= \frac{\Pr[M_1(X) = o^{(1)}]\Pr[M_2(X) = o^{(2)}]}{\Pr[M_1(X') = o^{(1)}]\Pr[M_2(X') = o^{(2)}]}\\
%     &\leq  \frac{e^{\epsilon_1}\Pr[M_1(X') = o^{(1)}] \cdot e^{\epsilon_2}\Pr[M_2(X') = o^{(2)}]}{\Pr[M_1(X') = o^{(1)}]\Pr[M_2(X') = o^{(2)}]} \\
%     &= e^{\epsilon_1 + \epsilon_2},
%   \end{align}
% 
%   and the Lemma follows. \Arielnote{What is that? what does it proof in the main paper?}
% \end{proof}

%\setcounter{theorem}{0}
%\setcounter{lemma}{0}

\begin{algorithm}[ht]
  \caption{DimReduction\\
    \textbf{Input:} $(x_1, x_2, \ldots, x_n), d, d', \beta$ \\
    \textbf{Output:} $(x'_1, \ldots, x'_n)$ low-dimensional space dataset.} 
  \label{alg:dimension-reduction}
  \begin{algorithmic}[1]
    \STATE $\Lambda = \sqrt{\frac{0.01 d}{\log(n/\beta)d'}}$
    \FOR {$i \in \{1,..,n\}$}
    \STATE $\tilde{x}_i \gets \Pi_{\mathcal{S}}(x_i)$
    \IF {$\Vert \tilde{x}_i \Vert \leq 1/\Lambda$} 
    \STATE $x'_i = \Lambda\tilde{x}_i$
    \ELSE
    \STATE $x'_i=0$
    \ENDIF
    \ENDFOR
    \STATE \textbf{return} $(x'_1, \ldots, x'_n)$
  \end{algorithmic}
\end{algorithm}
\vspace{-0.1in}
\begin{algorithm}[ht]
  \caption{DimReverse\\
    \textbf{Input:} $(c'_1, \ldots, c'_k), (x'_1, \ldots, x'_n)$ \\
    \textbf{Output:} $(c_1, \ldots, c_k)$ Private Centroids in high dimension
  }
  \label{alg:dimension-reverse}
  \begin{algorithmic}[1]
    \STATE $\X_1, \ldots, \X_k \gets \text{ the partition induced by }$  $(c'_1, \ldots, c'_k)$  on $(x'_1, \ldots, x'_n)$
    \FOR {$j\in \{1, \ldots, k\}$}
    \STATE $c_j\gets \textsc{FindCenter}^{\epsilon/2}(\X_j)$
    \ENDFOR
    \STATE \textbf{return} $(c_1, \ldots, c_k)$
  \end{algorithmic}
\end{algorithm}

\section{Additional proofs for \algo{}}

% \firstTherorem* 

% \begin{theorem}
% \label{theorem:privacy-full}
% (Full version of Theorem~\ref{theorem:privacy})
% The solution $c$ and $\cost{(S_{\epsilon})}$ released by Algorithm~\ref{alg:Private-Clustering} are $\epsilon$-DP.
% For all clients $i$, the value $\cost(\Se) - \cost{\Sie}$ released by Algorithm~\ref{alg:dp-explain} is $\epsilon$-$x_i$-exclusion DP.
% \end{theorem}

\mainprivacy*

\begin{proof}
It follows that $\cost(\Se)$ is the direct results of $Y$, which is $\epsilon/2$-differentially private coreset. By the post-processing property, $\cost(\Se)$ is $\epsilon/2$-DP (which implies $\epsilon$-DP).

The output $c$ of the \textsc{DimReverse} algorithm is $\epsilon/2$-differentially private with respect to the input $(X_1, X_2, \ldots, X_k)$, where $X_i$ represents the data points in cluster $i$. The overall process achieves $\epsilon$-differential privacy through composition, as $(X_1, X_2, \ldots, X_k)$ is partially derived from $Y$, which itself is $\epsilon/2$-differentially private.

%\Arielnote{I don't understand the notation }
For each explanations $\Sie$, let $X,X': X\setminus\{x_i\} \sim X'\setminus\{x_i\}$, i.e., $X$ and $X'$ are any two neighbor datasets that differ at exact one data point that is not agent $i$. Let $\Sie(Y^X)$ be the value of $\Sie$ with input dataset $X$ (and $Y^X$ as the private coreset of $X$ respectively). Note that we specify the dataset $X$ (and $X'$) as a parameter of $\Sie$ to highlight the original dataset where the explanation comes from (either $X$ or $X'$). Fix any set $S$, let $T = \{Y: \Sie(Y) \in S\}$ ,i.e., the set of coresets $Y$ that make $\Sie(Y) \in S$. Since $X\sim X'$, we have:
\begin{align}
  \Pr[\Sie(Y^X)\in S] &= \Pr[Y^X \in T] \\
                        &\leq e^{\epsilon/2}\Pr[Y^{X'}\in T]\\
  &= e^{\epsilon/2}\Pr[\Sie(Y^{X'})\in S],
\end{align}

which implies that $\cost(\Sie) - \cost(\Se)$ is $\epsilon$-$x_i$-exclusion DP, since $\cost(\Se)$ is $\epsilon/2$-DP (which implies $\epsilon/2$-$x_i$-exclusion DP).
\end{proof}

\validExplanation*

\begin{proof}
By the result of Lemma~\ref{lemma:JL}, with probability $1-2\beta$ we have:

\begin{align}
    cost(S^{(i)}_\epsilon) &\geq OPT_i \\ 
    &\geq w''(1+\alpha)OPT + t^{(i)}\\
    &\geq w''(1+\alpha)\frac{cost(S_\epsilon) -  \Omega_{p, \alpha, w''}\left(\frac{(k/\beta)^{O_{p,\alpha}(1)}}{\epsilon}.\plog(n/\beta)\right)}{w''(1+\alpha)} + t^{(i)} \\
    &= cost(S_\epsilon) + t^{(i)}  - \Omega_{p, \alpha, w''}\left(\frac{(k/\beta)^{O_{p,\alpha}(1)}}{\epsilon}.\plog(n/\beta)\right).
\end{align}

Set $t^{(i)} = \Omega_{p, \alpha, w''}\left(\frac{(k/\beta)^{O_{p,\alpha}(1)}}{\epsilon}.\plog(n/\beta)\right)$ and the Lemma follows.
\end{proof}

\begin{definition}
  \label{def:lambda}
  For $p \geq 1, \alpha > 0$, $\lambda_{p,\alpha/2} \overset{def}{=} \frac{1+\alpha/2}{((1+\alpha/2)^{1/p}-1)^p}$.
\end{definition}

\begin{lemma}
\label{lemma:JL}
  (Johnson-Lindenstrauss (JL) Lemma~\cite{johnson1984extensions,dasgupta2003elementary})
  Let $v$ be any $d$-dimensional vector. Let $\mathcal{S}$ denote a random $d'$-dimensional subspace of $\R^d$ and let $\Pi_S$ denote the projection from $\R^d$ to $\mathcal{S}$. Then, for any $\tau \in(0,1)$ we have

  \begin{align}
    \Pr\left[ \Vert v \Vert_2 \approx_{1+\tau}\sqrt{d/d'}\Vert \Pi_{\mathcal{S}}(v) \Vert_2  \right] \geq 1 - 2\exp\left(-\frac{d'\tau^2}{100} \right)
  \end{align}
\end{lemma}

\begin{lemma}
  \label{lemma:dimension-reduction} (Dimensionality Reduction for $(k,p)$-Cluster~\cite{makarychev2019performance})
  For every $\beta > 0, \tilde{\alpha} <1, p \geq 1, k\in\mathbb{N}$ , there exists $d' = O_{\tilde{\alpha}}(p^4\log(k/\beta))$. Let $\mathcal{S}$ be a random d-dimensional subspace of $\mathbb{R}^d$ and $\Pi_{\mathcal{S}}$ denote the projection from $\R^d$ to $\mathcal{S}$. With probability $1-\beta$, the following holds for every partition $\mathcal{X} = (X_1, \ldots, X_k)$ of $X$:

  \begin{align}
    cost^p(\mathcal{X}) \approx_{1+\tilde{\alpha}}(d/d')^{p/2}cost^p(\Pi_{\mathcal{S}}(\mathcal{X})),
  \end{align}

  where $\Pi_{\mathcal{S}}(\mathcal{X})$ denotes the partition $(\Pi_{\mathcal{S}}(X_1), \ldots,\Pi_{\mathcal{S}}(X_k) )$.
\end{lemma}

\explUtility*
%\begin{theorem}
%  (Full version of Theorem~\ref{theorem:explanation-utility})
%  Fix an $i$. With probability at least $1-\beta$, cost($S^{(i)}_\epsilon$) released by Algorithm~\ref{alg:dp-explain} is a $(w, t)$-approximation of the optimal cost $OPT_i$, in which:
%
%  \begin{align*}
%    w &= w'(1+\alpha)\\
%    t &= w'\bigO_{k,p}\left( \frac{2^{O_{p,\alpha}(d)}k^2\log^2{n}}{\epsilon}.\plog( \frac{n}{\beta} \right) ,  
%  \end{align*}
%  and $OPT_i$ is the optimal $(k,p)$-clustering cost with a center is fixed at position $x_i$.
%  \label{theorem:explanation-utility-full}
%\end{theorem}

\begin{proof}

  %\dungnote{This proof can be optimized to eliminate the component $d/d'$ by incorporating the $\Lambda$ part, I will do later}

  %\Arielnote{I'm not familiar with X tilda notation. And d tilda notation. }

  Let $\tilde{\X} = (\tilde{x}_1, \tilde{x}_2, \ldots, \tilde{x}_n)$, i.e., the projected data after applying the transformation $\Pi_{\mathcal{S}}$.
  Let $X = (x'_1, x'_2, \ldots, x'_n)$, i.e., $\tilde{\X}$ after being clipped by $1/\Lambda$.
  By setting $\alpha' = 0.1\alpha$, and applying Lemma~\ref{lemma:dimension-reduction}, we have:

  \begin{align}
  {OPT}_i^{\tilde{d}} \leq \left(\frac{d'}{d}\right)^{p/2}(1+0.1\alpha)OPT_i^{d}.
  \end{align}
  
  By standard concentration, it can be proved that $\Vert x'_i\Vert \leq 1/\Lambda$ with probability $0.1\beta/n$ as follows:

  Using Lemma~\ref{lemma:JL}, we have:

  \begin{align}
    \Pr\left[\Vert x \Vert > \frac{1}{1+\tau}\sqrt{d/d'}\Vert x'\Vert  \right] &\geq 1 - 2\exp(-d'\tau^2/100).
  \end{align}

  Since $x$ is in the unit ball, $\Vert x \Vert < 1$, which leads to:

  \begin{align}
    \Pr\left[\Vert x' \Vert < (1+\tau)\sqrt{d'/d}\right] &\geq 1 - 2\exp(-d'\tau^2/100).
  \end{align} 
    %\Arielnote{$\zeta = \sqrt{100 \cdot \frac{d^2}{d'^2} \cdot \log\left(\frac{n}{B}\right)} - 1$}
  Setting $\tau = \sqrt{\frac{log(n/\beta)}{0.01}} - 1$, $\Lambda = \frac{1}{1+\zeta}\sqrt{d/d'} = \sqrt{\frac{0.01}{\log{(n/\beta)}} . \frac{d}{d'}}$, we have:

  \begin{align}
    \Pr[\Vert x' \Vert < 1/\Lambda] &\geq 1 - 2\exp(-\frac{d'\tau^2}{100})\\
                       &> 1 - 2\exp(-d'\log(n/\beta))\\
                       %&> 1 - xp(-d'\log(n/\beta))\\
    &> 1 - 2\beta/n, 
  \end{align}%\dungnote{I missed the constant 0.1. Can any check where I missed it?}

  By union bound on all $i$, then with probability at least $1-2\beta$, $x'_i = \Lambda\tilde{x}_i$ for all $i$. Since $Y$ is the output of $\textsc{PrivateCoreset}$ with input $X'$ and $\zeta$, then by Theorem 38 of~\cite{ghazi:neurips20}, $Y$ is a $(0.1\alpha, t)$-coreset of $X'$ (with probability at least $1-\beta$), with $\alpha = \frac{(100\zeta)^p}{10\lambda_{p,\alpha/2}}$ and $t'$ as:

  \begin{align}
    t' = O_{p,\alpha}\left( \frac{2^{O_{p,\alpha}(d')}k^2\log^2{n}}{\epsilon}\log\left( \frac{n}{\beta} \right)  + 1\right).
  \end{align}

  We note that alternatively, given a target approximation parameter $\alpha$, we can set $\zeta = 0.01\left(\frac{\alpha}{10\lambda_{p,\alpha/2}}\right)^{1/p}$.

  Let $(y_1, y_2, \ldots, y_k)$ be the solution of $\textsc{NonPrivateApproxFC}$ in \algo{} for a fixed $i$, $(y^*_1, y^*_2, \ldots, y^*_k)$ be the optimal solution of the clustering with fixed center at $x'_i$ on $X$, $OPT_Y$ be the optimal cost of the clustering with fixed center at $x'_i$ on $Y$. By the $w'$-approximation property of $\textsc{NonPrivateApproximationFC}$, we have:

  \begin{align}
    cost_Y(y_1, y_2, \ldots, y_k) &\leq w' OPT_Y \\
                                  &\leq w' cost_Y(y^*_1, y^*_2, \ldots, y^*_k) \\
                                  &\leq w'(1+0.1\alpha) cost_{\X'_{1..n}}(y^*_1, y^*_2, \ldots, y^*_k) + w't' \\
    &= w'(1+0.1\alpha)OPT^{d'}_i + w't'.
  \end{align}

%  Also, we can apply the same argument (the $w$-approximation property of $\textsc{NonPrivateApproximationWithFixedCenter}$) to have:
%
%  \begin{align}
%    cost_{X'}(y_1, y_2, \ldots, y_k) &\leq \frac{1}{1-0.1\alpha}(cost_Y(y_1, y_2, \ldots, y_k) + t) \\
%                                     &\leq \frac{1}{1-0.1\alpha}(w'(1+0.1\alpha)OPT^{d'}_i + w't + t) \\
%    &\leq w'(1+\alpha)OPT^{d'}_i + O_{w'}(t).
%  \end{align}

  Composing with Lemma~\ref{lemma:dimension-reduction}, we have:

  \begin{align}
    cost_Y(y_1, y_2, \ldots, y_k) &\leq w'(1+0.1\alpha)OPT^{d'}_i + w't' \\
    &\leq \Lambda^{p}w'(1+0.1\alpha)OPT^{\tilde{d}}_i + w't'\\
                                  &\leq \Lambda^{p}w'(1+0.1\alpha)(1+0.1\alpha)\left(\frac{d'}{d}\right)^{p/2}OPT^d_i + w't'\\
    &\leq w'(1+\alpha)OPT^d_i\left(\frac{0.01}{\log(n/\beta)}\right)^{p/2} + w't', \mbox{ since $\Lambda^2d'/d = \Theta(1/\log(n/\beta))$}
  \end{align}

  Finally, since
  \begin{align}
    cost(S^{(i)}_\epsilon) &= cost_Y(y_1, y_2, \ldots, y_k)\left(\frac{\log(n/\beta)}{0.01}\right)^{p/2}\\
        &\leq w'(1+\alpha)OPT^d_i + \Theta(w't'(\log(n/\beta)^{p/2}) \\
        &\overset{(a)}{\leq} w'(1+\alpha)OPT^d_i + w'\Theta\left(\frac{2^{O_{p,\alpha}(d')}k^2\log^2{n}}{\epsilon}\log\left( \frac{n}{\beta} \right) (\log(n/\beta))^{p/2}\right) \\
        &\overset{(b)}{\leq} w'(1+\alpha)OPT^d_i + w'\Theta\left(\frac{(k/\beta)^{O_{p,\alpha}(1)}k^2\log^2{n}}{\epsilon}\log\left( \frac{n}{\beta} \right) (\log(n/\beta))^{p/2}\right) \\
        &\overset{(c)}{\leq} w'(1+\alpha)OPT^d_i + w'\Theta\left(\frac{(k/\beta)^{O_{p,\alpha}(1)}(k/\beta)^2\log^2(n/\beta)}{\epsilon}\log\left( \frac{n}{\beta} \right) (\log(n/\beta))^{p/2}\right) \\
        &= w'(1+\alpha)OPT^d_i + w'\Theta\left(\frac{(k/\beta)^{O_{p,\alpha}(1)}}{\epsilon}.\plog(n/\beta)\right), \\
  \end{align}

    where in $(a)$ we substitute the value of $t'$, and $(b)$ is because $d' = O_\alpha(p^4\log(n/\beta))$, and $(c)$ is because $\beta < 1$, and the Lemma follows.

\end{proof}

% \begin{lemma}
%   (Lemma~\ref{lemma:valid-explanation})
%   Fix an $i$. If $OPT_i \geq w''(1+\alpha)OPT + t^{(i)}$, then with probability at least $1-2\beta$, $f_0$ and $f_i$ released by Algorithm~\ref{alg:dp-low-dim-clustering} satisfies that $f_i > f_0$
%   \label{lemma:valid-explanation-full}
% \end{lemma}

% \begin{proof}
% By the result of Lemma~\ref{lemma:JL}, with probability $1-2\beta$ we have:

% \begin{align}
%     f_i &\geq OPT_i \\ 
%     &\geq w''(1+\alpha)OPT + t^{(i)}\\
%     &\geq w''(1+\alpha)\frac{f_0 -  \Omega_{p, \alpha, w''}\left(\frac{(k/\beta)^{O_{p,\alpha}(1)}}{\epsilon}.\plog(n/\beta)\right)}{w''(1+\alpha)} + t^{(i)} \\
%     &= f_0 + t^{(i)}  - \Omega_{p, \alpha, w''}\left(\frac{(k/\beta)^{O_{p,\alpha}(1)}}{\epsilon}.\plog(n/\beta)\right).
% \end{align}

% Set $t^{(i)} = \Omega_{p, \alpha, w''}\left(\frac{(k/\beta)^{O_{p,\alpha}(1)}}{\epsilon}.\plog(n/\beta)\right)$ and the Lemma follows.
% \end{proof}

\section{Additional proofs for $k$-median algorithm}

\begin{definition}
  \label{def:integer-program}
  The solution of the $k$-median problem (with demands and a center fixed at a location $z$ can be formulated as finding the optimal solution of the following Integer program (IP):

\begin{align}
    \text{minimize } &\sum_{i,j\in N} d_jc_{ij}x_{ij} \\
    \text{subject to } \sum_{i\in N} x_{ij} &= 1 \text{ for each $j\in N$}\\
    x_{ij} &\leq y_i \text{ for each $i, j\in N$}\\
    \sum_{j\in N} y_i &= k \\
    x_{ij} &\in \{0, 1\} \text{ for each $i, j\in N$}\\
    y_{i} &\in \{0, 1\} \text{ for each $i\in N$}\\
  y_z &= 1 \text{ for a fixed } z\in N \\
  x_{zz} &= 1 \text{ for a fixed } z \in N.
\end{align}

\end{definition}

%\anil{for all $i, j\in N$ in (28)}

\begin{lemma}
  \label{lemma:location-nearby}
    Locations $i, j \in N'$ satisfy: $c_{ij} > 4\max(\Cbar_i, \Cbar_j)$.
\end{lemma}

\begin{proof}
The lemma follows the demands moving step (in the first step of the algorithm): for every $j$ to the right of $i$ (which means $\Cbar_j \geq \Cbar_i$) and within the distance of $\Cbar_j$ (that also covers all points within distance $\Cbar_i$), we move all demands of $j$ to $i$, hence $j$ will not appear in $N'$. 
\end{proof}

% \begin{lemma}
%   \label{lemma:additive-cost}
% (Lemma~4 of~\cite{charikar1999constant}) 
% For any feasible integer solution $(x', y')$ for the input with modified demands, there is a feasible integer solution for the original input with cost at most $4\bar{C}_{LP}$ plus the cost of $(x', y')$ with demands $d'$.
% \end{lemma}

%\begin{lemma}
%    (Lemma~5 of~\cite{charikar1999constant})
%    For any feasible fractional solution $(\bar{x}, \bar{y})$, $\sum_{i: c_{ij} \leq 2\bar{C}_j}y_{i} \geq 1/2$ for each $j\in N$.
%\end{lemma}

% \begin{lemma}
%   \label{lemma:restricted-solution}
% (Theorem~6 of~\cite{charikar1999constant})
%     There is a $1/2$-restricted solution $(x', y')$ of cost at most $2\bar{C}_{LP}$.
% \end{lemma}

%\begin{lemma}
%  \label{lemma:minimum-cost}
%  (Lemma~7 of~\cite{charikar1999constant})
%    The minimum cost of $1/2$-restricted solution $(x', y')$ is $sum_{j\in N'}d'_j(1-y'_j)c_{s(j)j}$.
%\end{lemma}

\section{Additional proofs for $k$-means algorithm}

\begin{lemma}
  \label{lemma:centroidal}
  (Lemma 2.1 of~\cite{kanungo2002local}) Given a finite subset $S$ of points in $\mathbb{R}^d$, let $c$ be the centroid of $S$. Then for any $c' \in \mathbb{R}^d$, $\Delta(S, c') = \Delta(S, c) + |S|\Delta(c, c')$.
\end{lemma}

% \lemmaR*

%\begin{lemma}
%  (Full version of Lemma~\ref{lemma:R})
%  Let $S$ be $1$-stable set and $O$ be the optimal set of $k$ centers, we have $\Delta(O) - 3\Delta(S) + 2R \geq 0$, where $R = \sum_{q\in P}\Delta(q, s_{o_q})$.
%  \label{lemma:R-appendix}
%\end{lemma}

\begin{restatable}{lemma}{lemmaR}
  Let $S$ be $1$-stable set and $O$ be the optimal set of $k$ centers, we have $\Delta(O) - 3\Delta(S) + 2R \geq 0$, where $R = \sum_{q\in P}\Delta(q, s_{o_q})$.
  \label{lemma:R}
\end{restatable}

\begin{proof}
  Since $S$ is $1$-stable, we have for each swap pair:
  \begin{align}
    \sum_{q\in N_O(o)} &\big( \Delta(q, o) - \Delta(q, s_q) \big) \\
    + \sum_{q \in N_S(s) \setminus N_O(o)} &\big( \Delta(q, s_{o_q}) - \Delta(q, s)\big) \geq 0.
  \end{align}

  We will sum up the inequality above for all swap pairs. For the left term, the sum is overall $o \in O$:

  \begin{align}
    & \sum_{o \in O}\sum_{q\in N_O(o)} \big( \Delta(q, o) - \Delta(q, s_q) \big) \\
   = & \sum_{q \in P} \big( \Delta(q, o) - \Delta(q, s_q) \big),
  \end{align}

  Since each $o\in O$ will appear exactly once, and $\cup_{o\in O}q\in N_O(o)$ will cover all points in $P$.

  For the right term, the sum is over all $s$ that is being swapped out. We note that each $s$ can be swapped out at most twice, hence:

  \begin{align}
    \sum_{s \text{ being swapped out}}\sum_{q \in N_S(s)} &\big( \Delta(q, s_{o_q}) - \Delta(q, s)\big) \\
    \leq 2\sum_{q\in P}&\big( \Delta(q, s_{o_q}) - \Delta(q, s)\big)
  \end{align}

  When we combine the two terms, we have:
  \begin{align}
    \sum_{q \in P} \big( \Delta(q, o) - \Delta(q, s_q) \big) + 2\sum_{q\in P}\big( \Delta(q, s_{o_q}) - \Delta(q, s)\big) &\geq 0 \\
    \sum_{q\in P}\Delta(q, o_q) - 3\sum_{q\in P}\Delta(s, s_q) + 2\sum_{q \in P}\Delta(q, s_{o_q}) &\geq 0 \\
    \Delta(O) - 3\Delta(S) + 2R &\geq 0,
  \end{align}

  and the Lemma follows.
\end{proof}

\begin{lemma}
  \label{lemma:dist} (Proof in Lemma 2.2 \& 2.3 of~\cite{kanungo2002local})
  Let $\alpha^2 = \frac{\Delta(S)}{\Delta(O)}$, we have $\sum_{q\in P}dist(q, o_q)dist(q, s_q) \leq \frac{\Delta(S)}{\alpha}$
\end{lemma}

% \begin{lemma}
%   \label{lemma:R2-appendix}
%   (Full version of Lemma~\ref{lemma:R2})
%   With $R$ and $\alpha$ defined as above: $R \leq 2\Delta(O) + (1 + 2/\alpha)\Delta(S)$.
% \end{lemma}

%\lemmaRR*

\begin{restatable}{lemma}{lemmaRR}
  \label{lemma:R2}
  With $R$ and $\alpha$ defined as above: $R \leq 2\Delta(O) + (1 + 2/\alpha)\Delta(S)$.
\end{restatable}

\begin{proof}
  By Lemma~\ref{lemma:R}, we have:
%\Arielnote{We are missing lemma in here.}
  \begin{align}
    R &= \sum_{q\in P} \Delta(q, s_{o_q}) \\
      &= \sum_{o\in O}\sum_{q \in N_O(o)} \Delta(q, s_o)\\
    &=\sum_{o\in O\setminus\sigma}\sum_{q \in N_O(o)} \Delta(q, s_o) + \sum_{q\in N_O(\sigma)}\Delta(q, \sigma)\\
    &=\sum_{o\in O\setminus\sigma}\Delta(N_O(o), s_o) + \Delta(N_O(\sigma), \sigma)\\
    &\overset{(a)}{=}\sum_{o\in O\setminus\sigma}\big(\Delta(N_O(o), o)+ |N_O(o)|\Delta(o, s_o)\big)+ \Delta(N_O(\sigma), \sigma)\\
      &= \sum_{o\in O\setminus\sigma}\sum_{q \in N_O(o)} \big(\Delta(q, o) + \Delta(o, s_o)\big) + \sum_{q\in N_O(\sigma)}\Delta(q, \sigma) + \Delta(\sigma, o_\sigma)\\
    &=\sum_{o\in O}\sum_{q \in N_O(o)} \big(\Delta(q, o) + \Delta(o, s_o)\big)\\
    &\overset{(b)}{=}\sum_{o\in O}\sum_{q \in N_O(o)} \big(\Delta(q, o) + \Delta(o, s_q)\big)\\
      &\leq\sum_{q \in P} \big(\Delta(q, o_q) + \Delta(o_q, s_q)\big)\\
   &\overset{(c)}{\leq} \Delta(O) + \sum_{q\in P}(dist(o_q, q) + dist(q, s_q))^2\\
    &=2\Delta(O) + \Delta(S) + 2\sum_{q\in P} dist(q, o_q)dist(q, s_q)\\
    &\overset{(d)}{\leq}2\Delta(O) + \Delta(S) + (2/\alpha)\Delta(S),
  \end{align}

  where $(a)$ is because Lemma~\ref{lemma:centroidal} applies for all $o\in O\setminus\sigma$, $(b)$ is because $\Delta(o, s_o) \leq \Delta(o, s_q)$, $(c)$ is because the triangle inequality applies for $\Delta(o_q, s_q)$ and $(d)$ is because of Lemma~\ref{lemma:dist} and the Lemma follows.
\end{proof}

\subsection{Swap pairs mapping}
\label{sec:kmean-swap-mapping}

In this section, we describe the swap pairs mapping scheme for the $k$-means with a fixed center algorithm. We adapt the scheme of~\cite{matouvsek2000approximate} to accommodate the fixed center. We discuss the modifications in Section~\ref{sec:kmeans}. Here we discuss the complete mapping scheme.

At the last iteration of the algorithm, we always have a candidate set of centers $S$ that is $1$-stable, i.e., no single feasible swap can decrease its cost. We then analyze some hypothetical swapping schemes, in which we try to swap a center $s\in S$ with an optimal center $o\in O$. We utilize the fact that such single swaps do not decrease the cost to create some relationships between $\Delta(S)$ and $\Delta(O)$--the optimal cost. Particularly, these relationships are stated in Lemma~\ref{lemma:R} and Lemma~\ref{lemma:R2}.

Let $\sigma$ be the fixed center. We note that $\sigma \in S$ and $\sigma \in O$. Let $s_o$ be the closest center in $S$ for an optimal center $o\in O$, which means $o$ is captured by $s_o$. It follows that $s_\sigma = \sigma$. A center $s\in S$ may capture no optimal center (we call it lonely). We partition both $S$ and $O$ into $S_1, \ldots, S_r$ and $O_1,\ldots, O_r$ that $|S_i| = |O_i|$ for all $i$.

We construct each pair of partitions $S_i, O_i$ as follows: let $s$ be a non-lonely center, $O_i = \{o\in O: s_o = s\}$, i.e., $O_i$ is the set of all optimal centers that are captured by $s$. Now, we compose $s$ with $|O_i|-1$ lonely centers (which are not partitioned into any group from $S$) to form $S_i$. It is clear that $|S_i| = |O_i| \geq 1$.

We then generate swap pairs for each pair of partitions $S_i, O_i$ by the following cases:

\begin{itemize}
\item $|S_i| = |O_i| = 1$: let $S_i = \{s\}, O_i = \{o\}$, generate a swap pair $\{s, o\}$.
\item $|S_i| = |O_i| = m > 1$: let $S_i = \{s, s_{1..m-1}\}$ in which $s_{1..m-1}$ are $m-1$ lonely centers, let $O_i=\{o_{1..m}\}$, generate $m-1$ swap pairs $\{s_j, o_j\}$ for $j = 1..m-1$. Also, we generate a swap pair of $\{s_1, o_m\}$. Please note that $s$ does not belong to any swap pair, each $o_j$ belongs to exactly one swap pair, and each $s_j$ belongs to at most two swap pairs.
\end{itemize}

We then guarantee the following $3$ properties of our swap pairs:

\begin{enumerate}
\item each $o\in O$ is swapped in exactly once
\item each $s\in S$ is swapped out at most twice
\item for each swap pair $\{s, o\}$, $s$ either captures only $o$, or $s$ is lonely (captures nothing).
\end{enumerate}

\subsection{ $\gamma$-approximate candidate center set for fixed-center $k$-means.}
\label{sec:kmean-centroid-set}

We describe how to generate a $\gamma$-approximate candidate center set for $k$-means with fixed center $\sigma$ for a dataset $X\subset \R^d$. From the result of~\cite{matouvsek2000approximate}, we create a set $C'$ which is a $\gamma$-approximation centroid set of $X$. We will prove that $C = C'\cup \{\sigma\}$ forms a $\gamma$-approximate candidate center set for $k$-means with fixed center $\sigma$.

\begin{definition}
    \label{def:tolerance-ball}
    Let $S\subset\R^d$ be a finite set with its centroid $c(S)$. A $\gamma$-tolerance ball of $S$ is the ball centered at $c(S)$ and has radius of $\frac{\gamma}{3}\rho(S)$.
\end{definition}

\begin{definition}
    \label{def:approximate-centroid-set}
    Let $X\subset\R^d$ be a finite set. A finite set $C'\in\R^d$ is a $\gamma$-approximation centroid set of $X$ if $C'$ intersects the $\gamma$-tolerance ball of each nonempty $S\subseteq X$.
\end{definition}

\begin{lemma}
  \label{lemma:C-prime}
  (Theorem 4.4 of~\cite{matouvsek2000approximate}) We can compute $C'$--a $\gamma$-approximation centroid set of $X$ that has size of $O(n\gamma^{-d}\log(1/\gamma))$ in time $O(n\log{n}+n\gamma^{-d}\log(1/\gamma))$.
\end{lemma}

\begin{theorem}
  \label{theorem:C}
   Let $C = C'\cup\{\sigma\}$, in which $C'$ is a $\gamma$-approximation centroid set computed as Lemma~\ref{lemma:C-prime}, then $C$ is a $\gamma$-approximate candidate center set for $k$-means with fixed center $\sigma$.
\end{theorem}

\begin{proof}
  Let $O = (O_1, O_2, \ldots, O_k)$ be the optimal clustering in which $O_1$ is the cluster whose center is $\sigma$ (we denote it as $O_\sigma$). For any $S\subset \R^d$, we define $cost_S(c) = \sum_{x\in S} \Vert x - c\Vert^2$  and $cost(S) = cost_S(c(S))$ in which $c(S)$ is the centroid of $S$.
  By Definition~\ref{def:approx-candidate-center}, we will prove that there exists a set $c_1, c_2, \ldots, c_k\subset C$ and $c_1 = \sigma$ such that $cost(c_1, c_2, \ldots, c_k) \leq (1+\gamma)cost(O)$. We adapt the analysis of~\cite{matouvsek2000approximate} for the special center $\sigma$--which is not a centroid as other centers in $k$-means.

  First, we analyze the optimal cost. For any cluster except $O_\sigma$, its center is also the centroid $c(O_i)$ of the cluster, while $O_\sigma$ must have center $\sigma$:

  \begin{align}
    cost(O) &= \sum_{x\in O_\sigma}\Vert x-\sigma\Vert^2 + \sum_{i=2..k}\sum_{x\in O_i}\Vert x - c(O_i)\Vert^2\\
            &=cost_{O\sigma}(\sigma) + \sum_{i=2..k}cost(O_i)
  \end{align}

  Now, we construct $\{c_1, \ldots, c_k\}$ as follows: setting $c_1 =\sigma$, for $i = 2..k$, $c_i\in C'$ is the candidate center that intersects the $\gamma$-tolerance ball of cluster $O_i$. For $O_\sigma$, $cost_{O_\sigma}(\sigma) = cost(O_\sigma)$. For other clusters, $cost_{O_i}(c_i)\leq(1+\gamma)cost(O_i)$ as below:
  \begin{align}
    cost_{O_i}(c_i) &= \sum_{x\in O_i}\Vert x-c_i\Vert^2\\
    &\leq \sum_{x\in O_i}(\Vert x-c(O_i)\Vert + \Vert c(O_i)-c_i \Vert)^2\\
    &=cost(O_i) + 2\Vert c_i - c(O_i) \Vert\sum_{x\in O_i}\Vert x - c(O_i)\Vert + |O_i|\Vert c_i -c(O_i)\Vert^2\\
                    &\leq cost(O_i) + 2\gamma/3\rho(O_i)\sqrt{|O_i|}\sqrt{cost(O_i)} + |O_i|(\gamma/3\rho(O_i))^2\\
                    &\leq cost(O_i) + (2/3)\gamma cost(O_i) + (\gamma^2/9)cost(O_i)\\
                    &\leq (1+\gamma/3)^2cost(O_i)\\
    &\leq (1+\gamma)cost(O_i).
  \end{align}

  Let $(S_1, S_2, \ldots, S_k)$ be the Voronoi partition with centers $(c_1, c_2, \ldots, c_k)$, i.e., $S_i$ are points in the Voronoi region of $c_i$ in the Voronoi diagram created by $c_1, \ldots, c_k$, we have:
  \begin{align}
    cost(c_1, c_2,\ldots,c_k) &= cost_{S_1}(\sigma) + \sum_{i=2..k}cost(S_i)\\
                              &\overset{(a)}{\leq} cost_{S_1}(\sigma) + \sum_{i=2..k}cost_{S_i}(c_i)\\
                              &\overset{(b)}{\leq} cost_{O_\sigma}(\sigma) + \sum_{i=2..k}cost_{O_i}(c_i)\\
                              &\overset{(c)}{\leq} cost_{O_\sigma}(\sigma) + (1+\gamma)\sum_{i=2..k}cost(O_i)\\
                              &\leq(1+\gamma)cost(O),
  \end{align}
  where $(a)$ is because $cost(S_i)$ implies its minimal cost for any center, $(b)$ is because $S_i$s are picked by Voronoi partition which minimies the cost over $k$ partitions of seletecd $k$ centers, and $(c)$ is because $cost_{O_i}(c_i)\leq(1+\gamma)cost(O_i)$ as we proved above, and the Theorem follows.
\end{proof}

%%% Local Variables:
%%% mode: latex
%%% TeX-master: "main"
%%% End:

\subsection{Additional details on experiments}

\subsection*{D.3.1 Datasets and Experimental Setup}
\begin{figure*}[htbp]
    \centering
    \includegraphics[width=\textwidth]{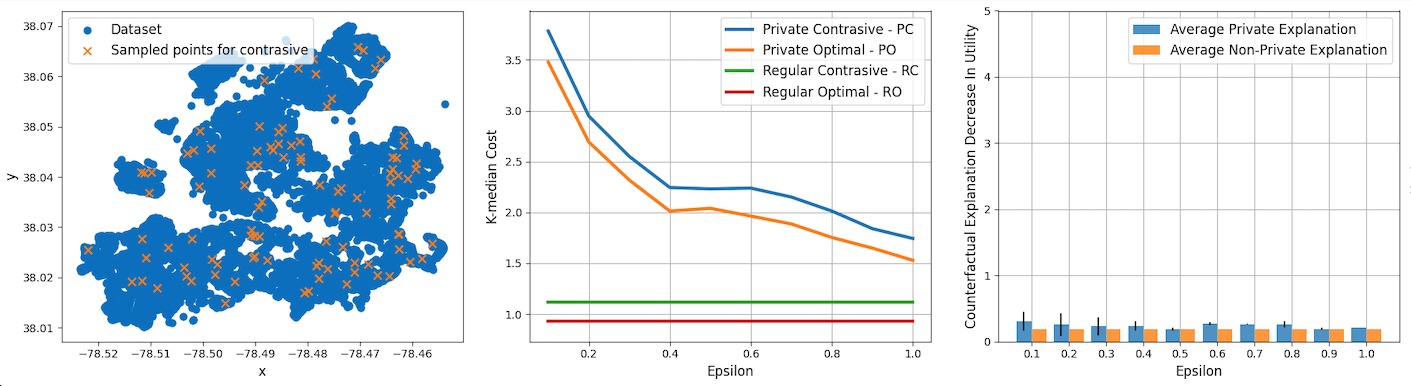}
    \caption{
        A detailed visualization of our dataset (Charlottesville County, Virginia) and analysis includes (a) A scatter plot of the full dataset with 100 randomly selected points for contrastive analysis, chosen to provide a more comfortable and manageable subset for explanation purposes. (b) Comparison of $k$-median clustering with fixed and non-fixed centroids, both private and non-private. (c) Bar graph showing contrastive explanation differences for differential private and non-private $k$-median with a fixed centroid.}
    \label{fig:CharlottesvilleUnited}
\end{figure*}

\begin{figure*}[htbp]
    \centering
    \includegraphics[width=\textwidth]{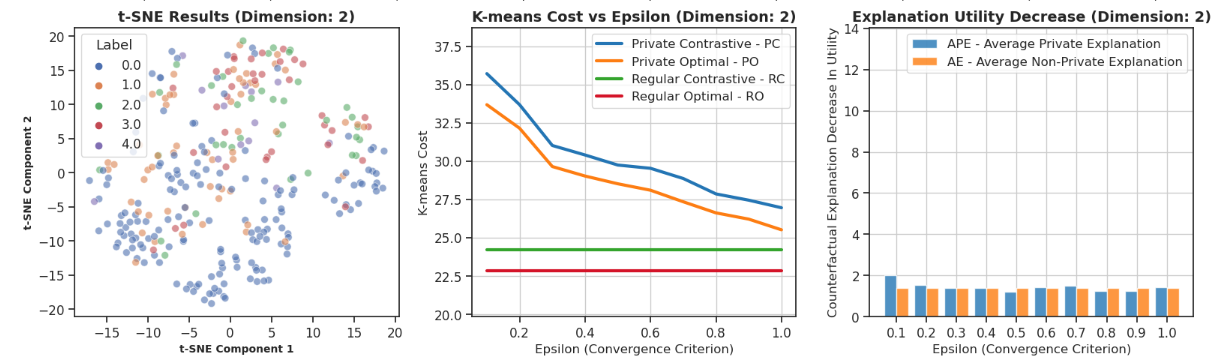}
    \includegraphics[width=\textwidth]{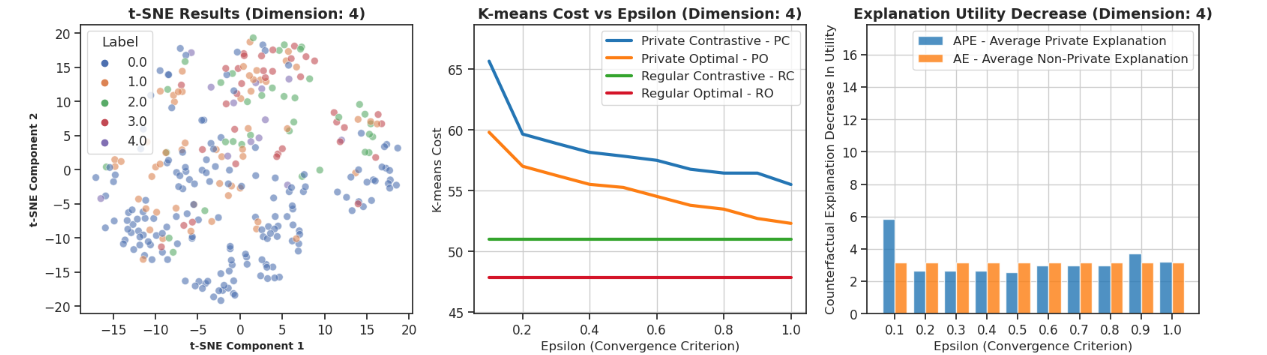}
    \includegraphics[width=\textwidth]{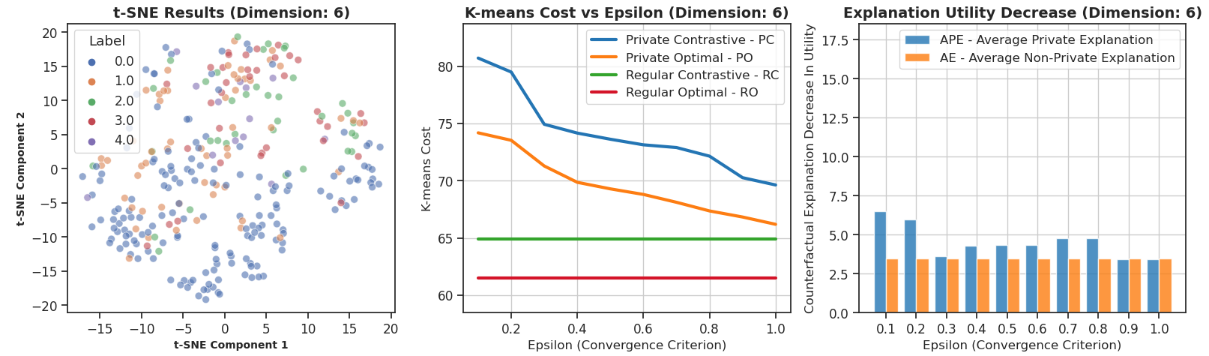}
    \includegraphics[width=\textwidth]{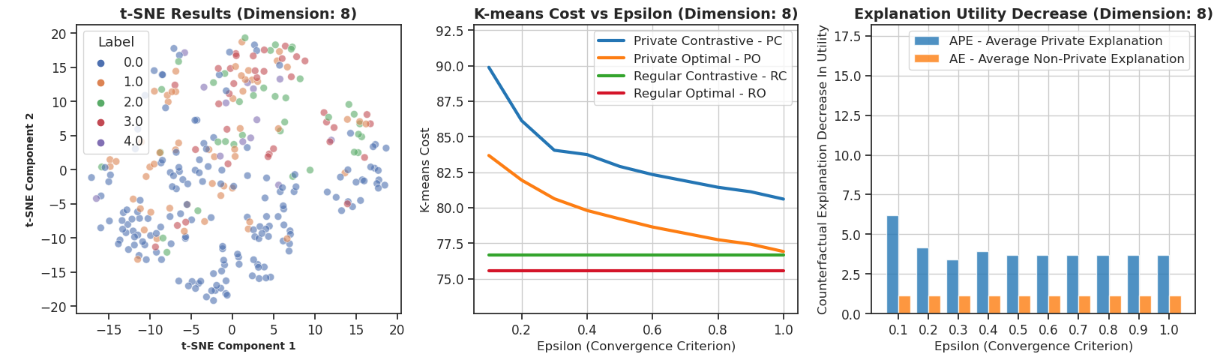}
        \caption{
        The figure presents visualizations of the dataset (Heart Disease, UCI MLR) reduced to four dimensions (2, 4, 6, and 8), along with the corresponding analysis (a) A t-SNE scatter plot illustrating the high-dimensional data. (b) Comparison of $k$-means clustering with fixed and non-fixed centroids, both private and non-private. (c) Bar graph showing contrastive explanation differences for differential private and non-private $k$-means with a fixed centroid.}
\end{figure*}

\begin{figure*}[htbp]
    \centering
    \includegraphics[width=\textwidth]{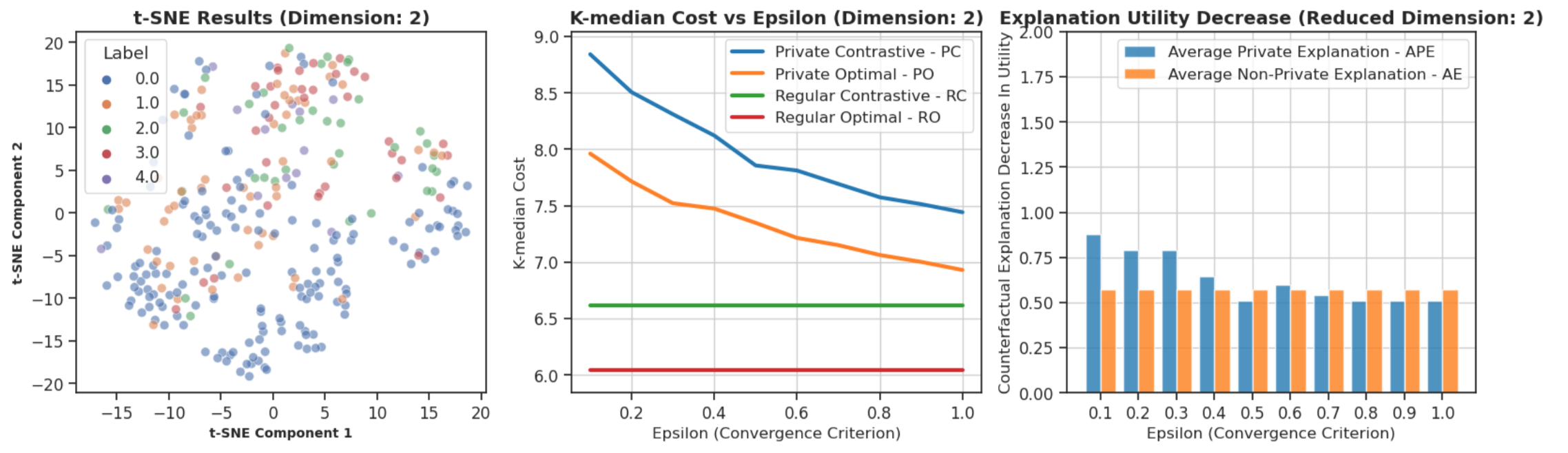}
    \includegraphics[width=\textwidth]{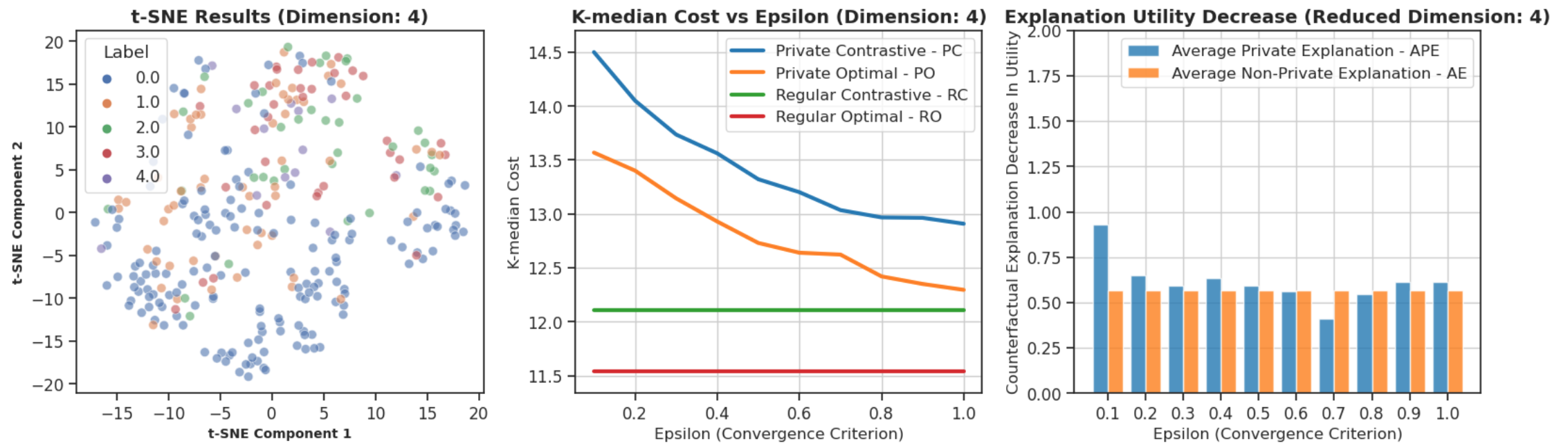}
    \includegraphics[width=\textwidth]{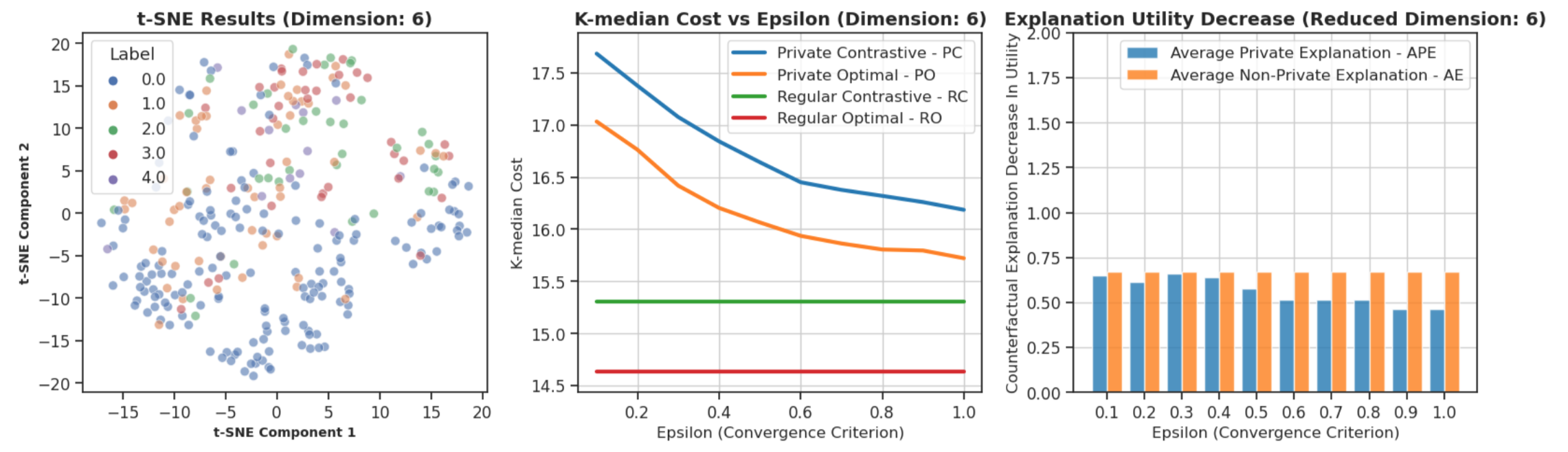}
    \includegraphics[width=\textwidth]{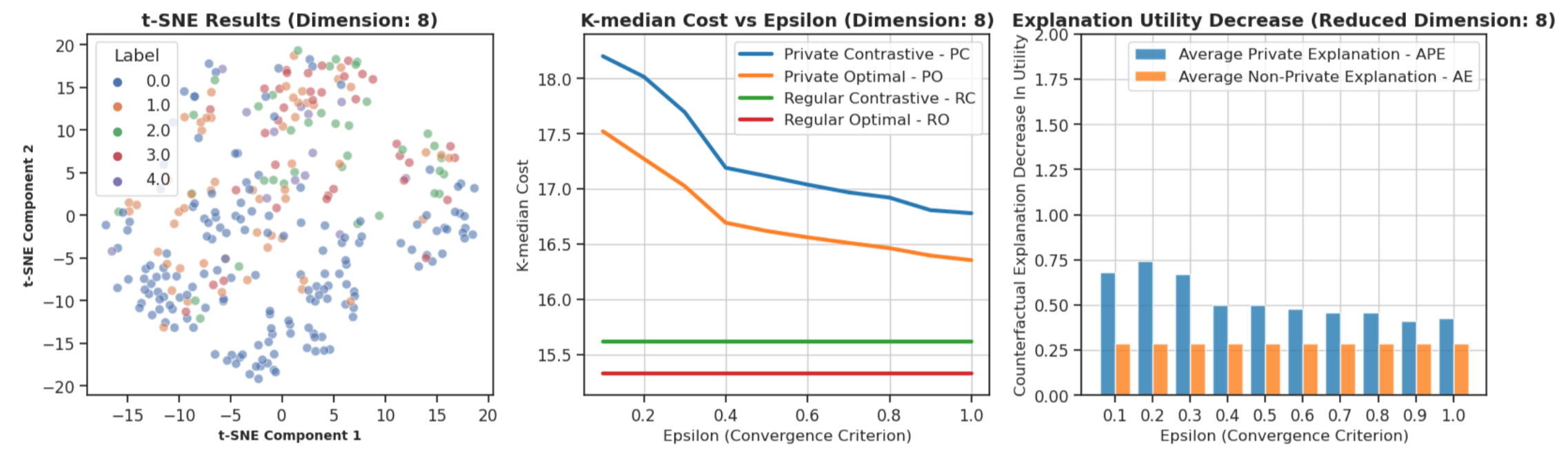}
        \caption{
        The figure presents further visualizations of the dataset (Heart Disease, UCI MLR) introduced in the main paper, reduced to four dimensions (2, 4, 6, and 8), along with the corresponding analysis (a) A t-SNE scatter plot illustrating the high-dimensional data. (b) Comparison of $k$-median clustering with fixed and non-fixed centroids, both private and non-private. (c) Bar graph showing contrastive explanation differences for differential private and non-private $k$-median with a fixed centroid.}
\end{figure*}

\begin{figure*}[htbp]
    \centering
    \includegraphics[width=\textwidth]{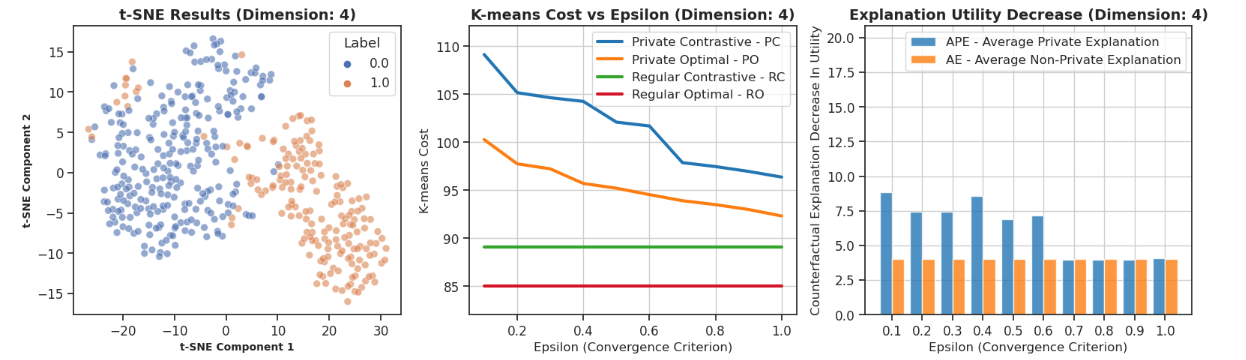}
    \includegraphics[width=\textwidth]{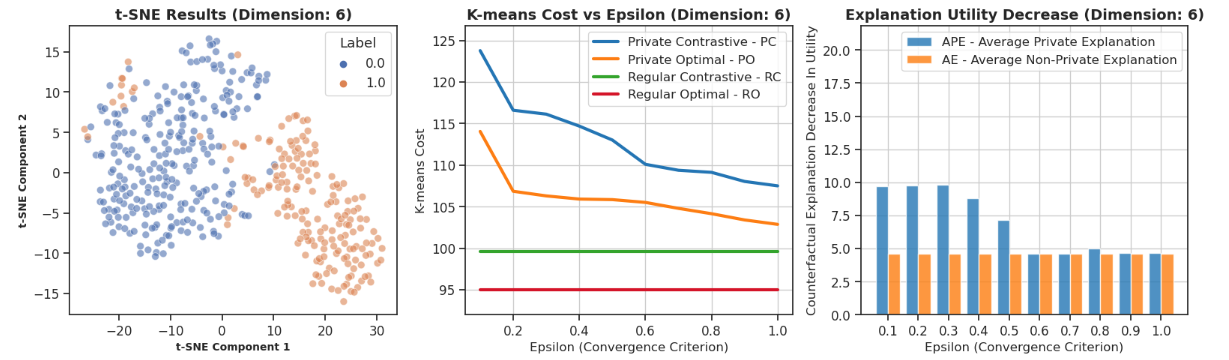}
    \includegraphics[width=\textwidth]{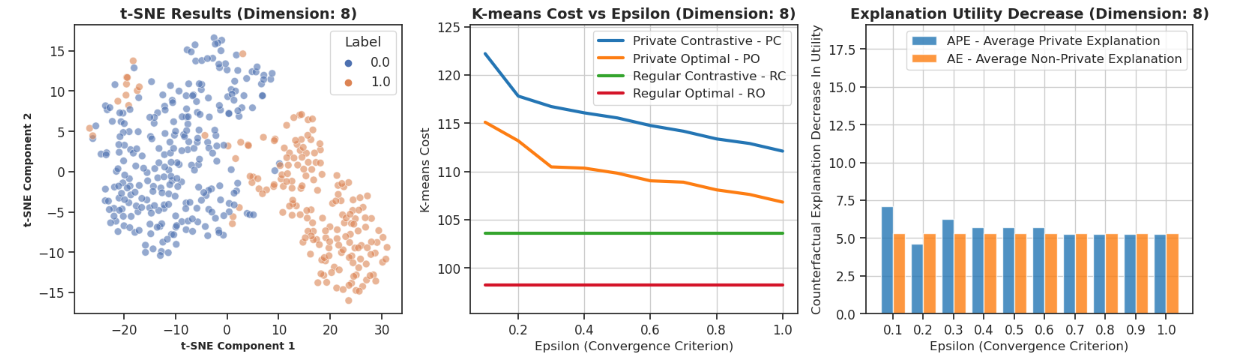}
    \includegraphics[width=\textwidth]{k-means/breast_cancer_data_kmenas/16.png}
        \caption{
        The figure presents visualizations of the dataset (Breast Cancer - 30 features, UCI MLR) reduced to four dimensions (4, 6, 8, and 16), along with the corresponding analysis (a) A t-SNE scatter plot illustrating the high-dimensional data. (b) Comparison of $k$-means clustering with fixed and non-fixed centroids, both private and non-private. (c) Bar graph showing contrastive explanation differences for differential private and non-private $k$-means with a fixed centroid.}
\end{figure*}

In the main body of the paper, we focused on the Heart Disease dataset from the UCI Machine Learning Repository, which was reduced from 13 features to 8 features for visualization purposes. However, we have also conducted experiments on the same dataset with different dimensionality reductions to analyze the impact on our results.
Furthermore, to demonstrate the robustness of our work, we have applied our analysis to additional datasets, including synthetic datasets and real-world datasets from Charlottesville, Albemarle and Breast Cancer Wisconsin dataset including 30 features. This appendix provides a detailed discussion of these experiments and their outcomes.
\subsection*{Real-World Datasets}
To demonstrate the robustness and real-world applicability of our approach, we conducted experiments on real-world datasets: the Charlottesville City dataset, the Albemarle County dataset, Heart Disease and Breast Cancer Wisconsin.

\subsection*{Charlottesville City Dataset}

The Charlottesville City dataset is part of the synthetic U.S. population, as described in [Chen et al., 2021] and [Barrett et al., 2009]. This dataset consists of approximately 33,000 individuals and around 5,600 activity locations visited by these individuals. The locations represent places where individuals perform various activities, providing insights into human mobility patterns and social interactions within the city.

\subsection*{Albemarle County Dataset}

The Albemarle County dataset is another real-world dataset used in our experiments. This dataset is significantly larger than the Charlottesville City dataset, comprising about 74,000 individuals. The increased sample size allows us to evaluate the scalability and performance of our approach when applied to larger, more complex datasets.

The Albemarle County dataset contains information about individuals' activities and the locations they visit, similar to the Charlottesville City dataset. This dataset provides a comprehensive representation of human mobility patterns and social interactions within the county.

By using these real-world datasets, we aim to validate the effectiveness and practicality of our methodology in real-life scenarios. The diverse characteristics of these datasets, such as the number of individuals and activity locations, enable us to assess the robustness and generalizability of our approach.

In our experiments, we applied our methodology to all datasets and compared the results to those obtained from the synthetic datasets. The consistency of results across these real-world datasets further reinforces the reliability and potential of our approach for real-world applications.
\subsubsection*{Synthetic 2D Dataset:}

We carefully created a synthetic dataset that mimics the properties of real-world datasets, striking a balance between realism and controlled variability. This dataset consists of 1000 uniformly distributed data points in a 2D space, with a range similar to the real datasets we analyzed.

The primary motivation behind this synthetic dataset is to provide a sandbox environment free from the unpredictable noise and anomalies of real-world data. This controlled setting is pivotal in understanding the core effects of differential privacy mechanisms, isolating them from external confounding factors. The dataset is a foundational tool in our experiments, allowing us to draw comparisons and validate our methodologies before applying them to more complex, real-world scenarios.

\begin{figure*}[htbp]
    \centering
    \includegraphics[width=1\textwidth]{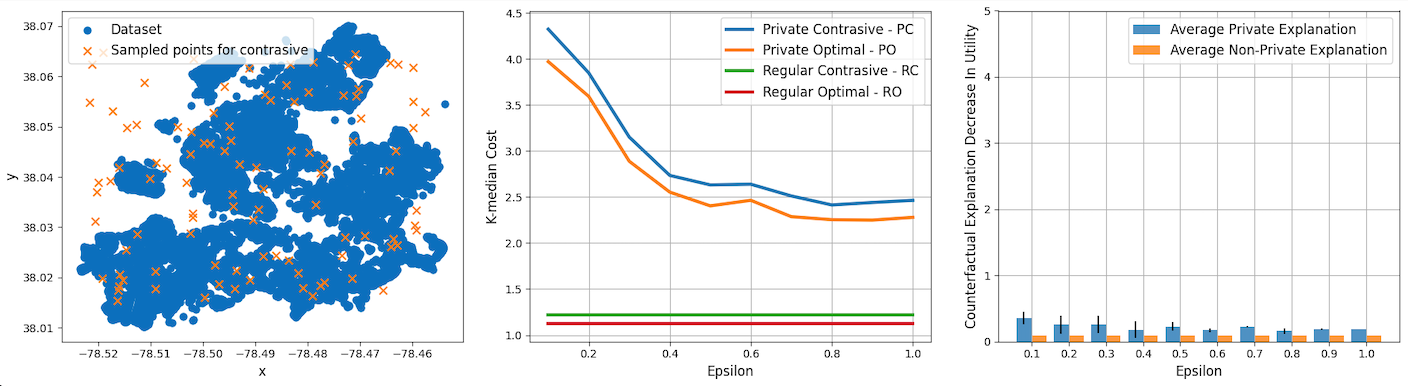}
    \caption{ A detailed visualization of our synthetic dataset, measured by the same metrics as the other real-world datasets.}
    \label{fig:synthetic}
\end{figure*}

\subsection*{D.3.2 Experimental Results}

\subsubsection*{Impact of $\epsilon$ on Private Optimal (PO) and Private Contrastive (PC):}
We observed consistent trends and patterns in all our data sets, including Charlottesville city, Albemarle county, and notably the Heart Disease dataset and the Breast Cancer dataset from the UCI Machine Learning Repository. As the value of $\epsilon$ increased to prioritize accuracy, we observed a gradual reduction in privacy protection. However, in line with our hypothesis, the impact of the epsilon budget on the explainability of our outcomes remained minimal. This consistency held true across all dimensions of the Heart Disease dataset, reinforcing the robustness of our findings across diverse data sources and attributes.

\subsection*{D.3.3 Performance Evaluation:}

For each $\epsilon$ value, we conducted 100 different runs for each dataset. The average results were consistent with our findings across all datasets. It's essential to note that these multiple invocations were solely for performance evaluation. In real-world applications, invoking private algorithms multiple times could degrade the privacy guarantee.
\subsubsection*{Consistency in Contrastive Explanations across Datasets:}

Despite the distinct scales between our different datasets, we observed consistent patterns in the contrastive explanations. Specifically, as illustrated in all Figures - (b), contrastive explanations remained largely unaffected by variations in the $\epsilon$ budget. This consistency further reinforces our hypothesis that the epsilon budget has a negligible influence on the explainability of our outcomes, even when applied to datasets of different scales.

\subsection*{D.3.4 Conclusion:}

The extended experiments on all our datasets further validate our approach's efficacy. The balance between privacy and utility, the robustness of contrastive explanations, and the negligible impact of $\epsilon$ on explainability were consistent across datasets. These findings underscore the potential of our method for diverse real-world applications.

% Add your bibliography file here
%\bibliography{sample}
%\bibliographystyle{plain}

\label{sec:expt-appendix}

%%% Local Variables:
%%% mode: latex
%%% TeX-master: "main"
%%% End:

% \clearpage

% \input{intro}

% \input{clustering}

\end{document}